\documentclass[final,twocolumn]{elsarticle}
\biboptions{numbers,sort&compress}

\usepackage{microtype} 
\usepackage{lmodern}
\usepackage[T1]{fontenc}
\usepackage{setspace}
\usepackage{booktabs}
\usepackage{longtable}
\usepackage{tabularx}
\usepackage{adjustbox} 
\usepackage{makecell}
\usepackage{ragged2e}
\usepackage{graphicx}
\usepackage{subcaption} 
\usepackage{float}
\usepackage{placeins}

\usepackage{amsmath}
\usepackage{amssymb}
\usepackage{bm}

\usepackage[hidelinks]{hyperref}
\usepackage{listings}

\usepackage{tikz}
\usetikzlibrary{calc}

\graphicspath{{figures/}{images/}}
\setkeys{Gin}{width=\linewidth,keepaspectratio}
\title{Portfolio Optimization Proxies under Label Scarcity and Regime Shifts --- Bayesian and Deterministic Students under Semi-Supervised Sandwich Training}

\date{}

\begin{document}

\begin{frontmatter}

\author[aff1]{Adhiraj Chattopadhyay\corref{cor1}}

\address[aff1]{Department of Management Studies, Indian Institute of Technology Roorkee, India}

\ead{adhiraj_c@ms.iitr.ac.in}
\ead{ufoundadhiraj@gmail.com}

\cortext[cor1]{Corresponding author}

\begin{abstract}
\noindent This paper proposes a machine learning assisted portfolio optimization framework designed for low data environments and regime uncertainty. We construct a teacher--student learning pipeline in which a Conditional Value at Risk (CVaR) optimizer generates supervisory labels, and neural models (Bayesian and deterministic) are trained using both real and synthetically augmented data.

The synthetic data is generated using a factor-based model with $t$-copula residuals, enabling training beyond the limited real sample of 104 labeled observations. We evaluate four student models under a structured experimental framework comprising (i) controlled synthetic experiments ($3\times5$ seed grid), (ii) in-distribution real-market evaluation (C2A), and (iii) cross-universe generalization (D2A).

In real-market settings, models are deployed using a rolling evaluation protocol where a frozen pre-trained model is periodically fine-tuned on recent observations and reset to its base state, ensuring stability while allowing limited adaptation.

Results show that student models can match or outperform the CVaR teacher in several settings, while achieving improved robustness under regime shifts and reduced turnover. These findings suggest that hybrid optimization--learning approaches can enhance portfolio construction in data constrained environments.
\end{abstract}

\end{frontmatter}


\section{Introduction}
\label{sec:introduction}

\subsection{The Portfolio Optimization Problem and Its Limitations}

Portfolio optimization is the allocation of capital across assets to balance
return and risk. The seminal work of Markowitz~\citep{markowitz1952} cast
the problem as a quadratic program trading expected returns against portfolio
variance, and subsequent work introduced tail-risk measures such as
Conditional Value-at-Risk (CVaR)~\citep{rockafellar2000} alongside
constraints accommodating turnover, leverage, and exposure limits.

Despite their theoretical elegance, classical approaches face persistent
practical limitations in live markets:
\begin{itemize}
    \item They assume normal distributions and linear inter-asset
    correlations, failing to account for the fat tails and nonlinear
    co-movement observed during market stress.
    \item Estimated expected returns and covariances are highly
    sensitive to small perturbations, causing large weight changes
    from minor forecasting errors.
    \item Optimizers solve each rebalance date independently, with
    no mechanism to generalize across regimes or adapt to
    structural market shifts.
    \item Trading constraints (turnover caps, position limits,
    transaction costs) are difficult to encode analytically without
    sacrificing convexity or tractability.
\end{itemize}
These issues produce a recurring gap between theoretically optimal
weights and realized out-of-sample performance.

\subsection{Machine Learning in Finance: Promise and Peril}

Machine learning offers a complementary direction. Neural models can
capture nonlinearities, time variation, and complex dependence patterns
that classical methods ignore~\citep{GuKellyXiu2020EAPML,
ChenPelgerZhu2021DLAP, Bagnara2024AssetPricingML}. However, naive
deep learning in portfolio construction introduces its own difficulties:
models overfit scarce and noisy data, behave unpredictably under regime
shifts, and violate financial constraints when training objectives are
not aligned with the allocation problem~\citep{feng2020, sawhney2021,
JiangXuLiang2017}.

The interpretability problem is more acute in finance than in other
domains. Classical models admit mechanistic explanations; deep networks
do not. When a model delivers strong performance for extended periods
and then fails, practitioners need diagnostics. Without uncertainty
quantification, there are none.

\subsection{Bayesian Neural Networks for Uncertainty-Aware Allocation}

Bayesian Neural Networks (BNNs) address this problem due to three structural properties:
\begin{itemize}
    \item They learn a posterior distribution over parameters rather than a single point estimate~\citep{Blundell2015BayesByBackprop, GalGhahramani2016}.
    This allows BNNs to capture predictive uncertainty directly. During periods of elevated uncertainty, posterior predictive dispersion increases and can be used to diagnose model unreliability under stress regimes.

    \item In data-constrained settings with many predictors (small-$n$, high-$p$; here $n = 104$, $p = 576$), standard neural networks can overfit and exhibit significant out-of-sample performance degradation. BNNs use variational inference as a regularizing mechanism, reducing overconfident parameter adaptation under limited evidence.~\citep{KingmaWelling2013AEVB}

    It should be noted that while we augment our dataset by generating synthetic data, it is drawn from the model of real data. It only adds structural coverage. It does not add genuinely new distributional information.

    \item Financial markets are not stationary. Regime mismatch can degrade model validity. Relative to deterministic networks, BNNs can signal distributional mismatch through elevated posterior predictive variance before large losses accumulate.
\end{itemize}

\subsection{Our Approach: Bayesian Knowledge Distillation via Sandwich
Training}

We build a portfolio allocation policy using the semi-supervised
sandwich training paradigm of~\citep{Pareek2025OptimizationProxies},
adapting it to financial optimization for the first time. A CVaR-minimizing
teacher generates allocation labels on labeled dates; a Bayesian student
learns to approximate and generalize the teacher's risk-aware behavior
via alternating supervised and unsupervised training phases:
\[
\parbox{\columnwidth}{\centering
\text{Teacher (CVaR optimizer)}\;$\rightarrow$\;
\text{Student (BNN, uncertainty-aware)}\\
$\rightarrow$\; \text{Real-data anchoring (partial labels)}.}
\]
Empirically, this produces three findings that motivate the design.
First, students consistently outperform the teacher, confirming that
distillation generalizes the \emph{structure} of CVaR optimization
rather than memorizing specific solutions. Second, Bayesian marginalization
produces implicit turnover regularization as an emergent property: without
any explicit turnover penalty in the training objective, BNN models
self-regulate to approximately half the weekly turnover of deterministic
counterparts. Third, learned policies transfer to a disjoint out-of-sample
asset universe with performance that \emph{improves} under high-volatility
conditions, through hierarchical decomposition of broad-market heuristics
into factor-level defensive positioning.

\subsection{Contributions}

This paper makes the following contributions:
\begin{enumerate}
    \item \textbf{A teacher–student learning framework for portfolio optimization.} We adapt the optimization-proxy paradigm for finance by using a CVaR (Conditional Value at Risk) optimizer, generating supervisory labels allowing the neural network to learn effectively even when dealing with sparse, volatile, and fat-tailed financial data.

    \item \textbf{A low-data BNN pipeline with natural uncertainty awareness.} We combine factor-based synthetic return generation with a Bayesian neural network. Unlike standard networks, the BNN provides calibrated uncertainty estimates. This prevents the model from over-committing when data is limited or market regimes shift.

    \item \textbf{Implicit turnover reduction without explicit constraints.} We adapt the optimization-proxy
    paradigm~\citep{Pareek2025OptimizationProxies} to portfolio
    construction --- to our knowledge the first such application in
    finance. We found a highly valuable emergent property of this Bayesian setup: the model naturally self-regulates its weekly trading turnover to $11$--$14\%$. This is approximately a $50\%$ reduction in trading activity compared to deterministic models, significantly lowering transaction costs without requiring hard-coded turnover penalties. Thereby we demonstrate that it generalizes effectively to
    non-stationary, fat-tailed financial returns under severe
    label scarcity.

    \item \textbf{Structured stress testing revealing hierarchical generalization.}
    Using a C2A and D2A testing protocol under controlled stress scenarios, we demonstrate the "HIGH-VOL paradox." Models trained only on broad, aggregated indices achieved a $+140\%$--$+276\%$ Sharpe ratio improvement during high-volatility regimes when tested on a completely new set of individual assets. They exhibited large HIGHVOL improvements in D2A. These  improvements suggest the transfer of broad risk-reduction heuristics to a more factor decomposed universe.
\end{enumerate}

The rest of this paper is organized as follows.
Section~\ref{sec:problem_statement} formalizes the portfolio
optimization problem and the learning objective.
Section~\ref{sec:related_work} situates our approach within the
literature on classical optimization, machine learning in finance,
and Bayesian deep learning.
Section~\ref{sec:data_preprocessing} describes the dataset
construction and synthetic augmentation pipeline.
Section~\ref{sec:methodology} presents the student architectures,
learning objectives, sandwich training schedule, and evaluation
design. Section~\ref{sec:experiments} reports experimental results
across all three evaluation tiers. Section~\ref{sec:discussion}
interprets the principal findings and discusses limitations.
Section~\ref{sec:future_work} outlines extensions, and
Section~\ref{sec:conclusion} concludes.

\section{Problem Statement}
\label{sec:problem_statement}

\subsection{Classical Portfolio Optimization}
\label{subsec:classical_portfolio}

Let $\mathbf{w} = [w_1, \dots, w_n]^\top \in \mathbb{R}^n$ denote portfolio
weights over $n$ risky assets, with $\mathbf{R} \in \mathbb{R}^n$ the vector
of random returns, $\boldsymbol{\mu} = \mathbb{E}[\mathbf{R}]$ expected
returns, and $\boldsymbol{\Sigma} = \mathrm{Cov}(\mathbf{R})$ the covariance
matrix. The classical Markowitz mean-variance problem
\citep{markowitz1952} is:
\begin{equation}
\min_{\mathbf{w}}\ \mathbf{w}^\top \boldsymbol{\Sigma} \mathbf{w}
\quad \text{s.t.}\quad
\mathbf{w}^\top \boldsymbol{\mu} \ge R^*,\quad
\mathbf{w}^\top \mathbf{1} = 1,\quad
\mathbf{w} \succeq \mathbf{0},
\label{eq:markowitz}
\end{equation}
where $R^*$ is a minimum acceptable return. This quadratic program yields
the efficient frontier but relies on stable second moments and penalizes
upside and downside risk symmetrically, making it ill-suited to fat-tailed,
regime-shifting markets.

\subsection{CVaR as the Risk Measure}
\label{subsec:cvar_formulation}

Conditional Value-at-Risk (CVaR) addresses the tail-risk limitations of
variance by directly controlling expected losses beyond a specified
confidence level \citep{rockafellar2000}. For portfolio loss
$L = -\mathbf{w}^\top \mathbf{R}$ and confidence level $\alpha$:
\begin{equation}
\mathrm{CVaR}_\alpha(\mathbf{w})
= \mathbb{E}\!\left[L \mid L \ge \mathrm{VaR}_\alpha(\mathbf{w})\right].
\label{eq:cvar_def}
\end{equation}
Following \citep{rockafellar2000}, minimizing CVaR admits a convex
reformulation with auxiliary variable $\ell$:
\begin{equation}
\begin{aligned}
\min_{\mathbf{w},\ell}\quad
&\ell + \frac{1}{(1-\alpha)T}\sum_{t=1}^{T}
\max\!\left\{-\mathbf{w}^\top \mathbf{R}_t - \ell,\ 0\right\}\\
\text{s.t.}\quad
&\mathbf{w}^\top \mathbf{1} = 1,\quad \mathbf{w} \succeq \mathbf{0},
\end{aligned}
\label{eq:cvar_ru}
\end{equation}
where $\mathbf{R}_t$ are historical or simulated scenario returns.
CVaR is convex, admits a sample-average objective, and is differentiable
almost everywhere, making it suitable both as a classical optimization
target and as a supervised training signal for neural approximators.
We use this formulation as the teacher objective throughout.

\subsection{From Optimization to Learning}
\label{subsec:learning_formulation}

Classical approaches --- whether mean-variance or CVaR-based --- face
four persistent limitations in practice: distributional mismatch (empirical
returns exhibit fat tails and nonstationarity), estimation fragility
(small errors in $\boldsymbol{\mu}$ and $\boldsymbol{\Sigma}$ propagate
into large weight changes), scenario blindness (optimizers do not
generalize to unseen stress regimes), and constraint rigidity (realistic
trading frictions are difficult to encode analytically).

Rather than solving the optimization problem from scratch at each rebalance
date, we learn a \emph{policy mapping} from market features to portfolio
weights:
\begin{equation}
f_\theta : \mathcal{I}_t \rightarrow \mathbf{w}_t,
\label{eq:policy_mapping_ps}
\end{equation}
where $\mathcal{I}_t$ is the information set available at time $t$
(historical returns, factor realizations, portfolio state) and $\theta$
are learned parameters. The objective is to learn $f_\theta$ such that
it generalizes across unseen regimes and asset universes, respects
portfolio constraints, and provides calibrated uncertainty estimates ---
properties that classical optimizers do not simultaneously satisfy.

\section{Related Work}
\label{sec:related_work}

Portfolio optimization has a long theoretical foundation rooted in the
mean-variance framework of \citep{markowitz1952}, subsequently extended
through equilibrium pricing \citep{sharpe1964, lintner1965}, the
Black--Litterman model \citep{black1992}, and the development of coherent
risk measures, most notably CVaR \citep{rockafellar2000}. CVaR optimization
retains convexity while directly controlling tail losses, making it
particularly attractive for applications subject to extreme events. Despite
their elegance, classical convex approaches depend on static distributional
assumptions and historical covariance estimates that break down in
non-stationary markets, motivating the integration of adaptive,
learning-based components.

The intersection of machine learning and asset pricing has grown
substantially in recent years. \citep{GuKellyXiu2020EAPML} demonstrate that
flexible nonlinear models can outperform linear factor models in
cross-sectional return prediction, while \citep{ChenPelgerZhu2021DLAP} and
\citep{Bagnara2024AssetPricingML} further establish the role of deep learning
in empirical asset pricing. Factor-based forecasting using Fama--French
models \citep{fama1993, fama2015} augmented with momentum \citep{carhart1997}
remains a standard and interpretable baseline; for Indian equity markets
specifically, \citep{AgarwallaJacobVarma2014IndiaFourFactor} provide an
adapted four-factor framework that informs our feature construction for the
Indian assets in our training universe. Direct portfolio construction using
neural architectures --- including recurrent networks, temporal convolutional
networks, and attention mechanisms \citep{feng2020, sawhney2021} --- and
reinforcement learning formulations \citep{moody2001, JiangXuLiang2017,
Liang2018AdversarialDRLPortfolio, LiuZhengCartlidge2025} have also been
explored, but these approaches typically sacrifice interpretability and
require extensive reward shaping, large training sets, and careful
hyperparameter tuning to achieve stable behavior under realistic constraints.

Probabilistic and Bayesian approaches offer a principled remedy to the
brittleness of point-estimate neural models. Bayesian neural networks
\citep{Blundell2015BayesByBackprop} replace fixed parameters with posterior
distributions, enabling uncertainty-aware predictions and improved
generalization under limited data. Scalable approximate inference via
variational inference and Monte Carlo dropout \citep{GalGhahramani2016,
KingmaWelling2013AEVB} has made BNNs practical for moderate-scale
applications. Bayesian perspectives have been applied to return forecasting
and regularization in asset pricing \citep{FengHePolson2018,
FengHePolsonXu2023, DixonPolsonGoicoechea2022}, but existing work focuses
on prediction rather than downstream portfolio construction, and few studies
explicitly incorporate portfolio-level constraints such as turnover limits,
box bounds, or tail-risk objectives into the learning problem.

The methodological foundation of our approach is the semi-supervised
sandwich training paradigm introduced by \citep{Pareek2025OptimizationProxies}
for power-grid optimization under limited labeled data. This framework
alternates between supervised imitation of scarce optimal solutions and
unsupervised penalty-based training that enforces known structural
constraints, learning an optimizer proxy rather than a direct decision policy.
The paradigm is especially attractive for financial optimization, where
optimal portfolios are expensive to compute and regime-dependent. Synthetic
data augmentation provides a complementary tool for expanding the labeled
training pool: \citep{Hoffmann2019DataLimitedML} demonstrate its value in
data-limited regimes, and copula-based dependence models
\citep{Patton2012CopulaReview, SalvatierraPatton2014, OhPatton2021} provide
a principled mechanism for preserving cross-asset tail dependence in
synthetic scenario generation, as used in our pipeline
(Section~\ref{subsec:synthetic_augmentation}).

To the best of our knowledge, no prior work applies Bayesian neural networks
with a semi-supervised sandwich training paradigm to portfolio optimization.
More broadly, existing evaluation protocols in this literature rely on fixed
asset universes and historical backtests, implicitly assuming stationarity.
Our experimental design departs from this norm by explicitly separating
training stability (GRID\_3$\times$5 synthetic evaluation), in-distribution
stress robustness (C2A) - \textbf{An evaluation on familiar assets under varied operational conditions or Constrained Application Assessment}, and out-of-distribution generalization to an unseen
asset universe (D2A) - \textbf{Cross Universe Adaptive Generalization}, providing a more stringent test of deployability under
realistic market uncertainty (Section~\ref{subsec:evaluation_design}).
To be more specific, we distinguish between two evaluation settings:

C2A:  
This setting evaluates model robustness within the same asset universe used during training. The evaluation is conducted on real market data under varying conditions, including stress scenarios and operational constraints. It measures in-distribution performance and sensitivity to market regimes.

D2A :  
This setting evaluates cross-universe generalization by applying trained models to a different set of assets not used during training. While the time period and feature construction remain consistent, the asset universe is partially disjoint (~$40\%$ overlap), ensuring that models cannot rely on memorization of asset-specific patterns.

Importantly, D2A is a cross-universe generalization test rather than a strictly out-of-time evaluation.

\section{Dataset and Preprocessing}
\label{sec:data_preprocessing}

This section describes the construction of the canonical dataset used for training and
evaluation. The full pipeline --- covering currency conversion, return cleaning rules,
factor normalization, feature equations, and synthetic generation mechanics --- is
detailed in ~\ref{app:data_details}.

\subsection{Asset Universe}
\label{subsec:universe}

We construct a weekly rebalanced, long-only portfolio over a universe 
of $N = 36$ assets. Weekly adjusted close prices were obtained via the \texttt{yfinance} 
library~\citep{yfinance} (Yahoo Finance), covering January~2015 through 
early~2026. Adjusted prices incorporate splits and dividend 
corrections as reported by Yahoo Finance. Given the weekly rebalancing 
frequency and the static historical nature of the backtest, this data 
source is appropriate for the research context. The assets are drawn from three groups: 23 US-listed ETFs (including 
broad equity, fixed income, commodity, and sector funds), 12 Indian 
equities converted to a USD base for currency consistency, and one 
market proxy. All prices are expressed in USD prior to return computation. This 
ensures a well-defined portfolio P\&L and a consistent risk model across 
the full universe. Weekly adjusted close prices are available from 
January 2015 through December 2025. All together, this yields a cleaned return panel 
of 575 weeks after alignment and coverage filtering.

ETFs are chosen over individual equities and raw indices for three 
interconnected reasons.

\textbf{Liquidity and execution fidelity.} ETFs trade continuously, 
have tight bid--ask spreads and have no minimum lot size. This ensures that the 
CVaR teacher can solve exactly at each rebalance date without 
microstructure interference. Individual equities introduce corporate 
event risk, earnings noise, and survivorship bias (assets active in 
2015 but delisted by 2025 would corrupt the learning signal in ways 
that are difficult to detect and correct). ETF closures are rare. The 
broad-market instruments used here have continuous history.

\textbf{Cross-asset diversification within a tractable universe.} 
The 36 ETFs span US equity, fixed income, commodities, real estate, 
and international markets. They provide genuine cross-asset coverage 
while keeping $N$ small enough for the CVaR teacher to remain 
computationally exact at each rebalance date. Individual equities 
at $N = 36$ would be poorly diversified within any single sector; 
broad indices at $N = 36$ would collapse the factor structure that 
the out-of-universe evaluation explicitly relies upon 
(Section~\ref{subsec:evaluation_design}).

\textbf{Retail deployment alignment.} As discussed in 
Section~\ref{sec:future_work},
the intended application of this framework is retail portfolio systems 
operating under realistic frictions. ETFs are the natural instrument 
for this context as they are low cost, tax efficient, and accessible at minimal 
capital without lot-size constraints that would otherwise break our 
weight model.

\subsection{Pipeline Overview}
\label{subsec:pipeline_overview}

The data pipeline proceeds in five stages. First, daily adjusted close prices are
resampled to a common weekly calendar (last observation per week) and INR-denominated
assets are converted to USD. Second, simple and log returns are computed and a
conservative missing-data policy is applied: assets with weekly coverage below 90\%
are removed and no zero-filling of returns is permitted, as this would distort tail
behavior under CVaR training \citep{rockafellar2000}. Third, weekly Fama--French
five-factor and momentum data \citep{fama1993, fama2015, carhart1997} are aligned to
the return panel and normalized to weekly decimals. Fourth, a 576-dimensional feature
vector is constructed for each rebalance date using only information available up to
that date (no look-ahead). Fifth, supervised labels are generated by solving a
scenario-based CVaR minimization problem at each eligible date
\citep{rockafellar2000, markowitz1952}, producing 104 real feature-label pairs after
a 104-week warm-up period.

\subsection{Feature Summary}
\label{subsec:feature_summary}

At each rebalance date $t$, we construct a per-asset feature matrix
$X_t \in \mathbb{R}^{N \times 16}$, flattened to a vector $x_t \in \mathbb{R}^{576}$.
The 16 features per asset are organized into six blocks as summarized in
Table~\ref{tab:feature_summary}. Expected returns are estimated via a rolling ridge
regression on the factor set, blended with a cross-sectional momentum signal. PCA
loadings are computed over a 104-week rolling window and capture dominant
cross-asset co-movement modes. Previous portfolio weights are included as context,
enabling the model to learn stable rebalancing transitions rather than independent
weight-by-weight decisions.

\begin{table}[htbp]
\centering
\caption{Feature blocks and their composition. Total: 16 features per asset,
576-dimensional flattened input vector for $N=36$ assets.}
\label{tab:feature_summary}
\begin{adjustbox}{max width=\columnwidth}
\begin{tabular}{llc}
\toprule
\textbf{Block} & \textbf{Features} & \textbf{Count} \\
\midrule
Return forecast / uncertainty & Blended expected return $\hat{\mu}^{\text{blend}}$,
    forecast uncertainty $\sigma_\mu$, realized volatility & 3 \\
PCA structure & Loadings on PC1, PC2, PC3 (rolling 104-week window) & 3 \\
Momentum & 12--1 momentum z-score, 6-month z-score, 1-month z-score & 3 \\
Drawdown & Rolling 52-week drawdown & 1 \\
State \& constraints & Previous weight $w_{i,t-1}$, position cap $c_i$ & 2 \\
Market regime & 4-week and 12-week market return, market volatility,
    market drawdown & 4 \\
\midrule
\textbf{Total} & & \textbf{16} \\
\bottomrule
\end{tabular}
\end{adjustbox}
\end{table}

\subsection{Synthetic Augmentation}
\label{subsec:synthetic_augmentation}

The 104 real labeled pairs represent a low-data regime: 576-dimensional inputs with
only $\sim$100 labeled samples is a classic small-$n$, high-$p$ setting
\citep{Hoffmann2019DataLimitedML}. To support learning under this constraint, we
augment training with synthetic market trajectories generated from an empirically
grounded factor-residual model. Factor dynamics are simulated via a VAR(1) process
fitted to real weekly Fama--French and momentum data, per-asset returns are
reconstructed from static Ridge factor loadings and idiosyncratic residuals, and
cross-asset tail dependence is preserved through a $t$-copula fitted to historical
residuals \citep{Patton2012CopulaReview}. Simulating 1,400 synthetic weeks and
applying the same feature builder and CVaR teacher to stride-sampled dates yields
approximately 323 additional labeled pairs. Full generation mechanics are given in
~\ref{app:synthetic_details}. Sanity checks are consistent with synthetic returns
exhibit slightly fatter tails than real data (median $|$return$|$ 1.94\% vs.\ 1.61\%)
and that the cross-asset correlation structure is preserved.

\subsection{Dataset Composition and Splits}
\label{subsec:dataset_splits}

The combined training pool consists of 104 real and 323 synthetic labeled pairs,
totaling 427 samples. The dataset is split 60/20/20 by construction: real pairs
occupy the training split in full, while validation and test rows are drawn from the
synthetic pool. This is a deliberate design choice --- real labeled observations are
scarce and structurally informative, and are therefore retained entirely for
supervised training. The 323-sample synthetic pool also provides the unlabeled
scenario windows used during the unsupervised phases of sandwich training.

It is important to note that the real data spans 2015--2026, and the training split
therefore includes observations up to approximately 2024. The C2A evaluation
(Section~\ref{subsec:evaluation_design}) applies frozen models to the same
36-asset universe over the 2022--2026 period and should be interpreted as an
in-distribution stress and constraint sensitivity analysis rather than out-of-sample
generalization. True out-of-sample generalization is assessed exclusively through
the D2A evaluation on a disjoint 36-asset universe, described in
Section~\ref{subsec:evaluation_design}.

\section{Methodology}
\label{sec:methodology}

\subsection{Problem Setup and Notation}
\label{subsec:problem_setup}

We consider a weekly rebalanced, long-only portfolio allocation problem over a
universe of $N=36$ tradable assets. Let $t$ index discrete weekly rebalance dates
and let $\mathbf{r}_t \in \mathbb{R}^{N}$ denote the vector of simple asset returns
at week $t$. Portfolio weights chosen at decision time $t$ are denoted
\begin{equation}
\mathbf{w}_t \in \mathbb{R}^{N}, \quad \mathbf{w}_t \succeq \mathbf{0},
\quad \mathbf{1}^\top \mathbf{w}_t = 1,
\label{eq:weights_simplex}
\end{equation}
so weights lie on the probability simplex \citep{markowitz1952}. Under the
anti-leakage convention, weights chosen at $t$ are applied to returns at $t+1$:
\begin{equation}
r^{p}_{t+1} = \mathbf{w}_t^\top \mathbf{r}_{t+1}.
\label{eq:portfolio_return}
\end{equation}
Given the 576-dimensional feature vector $\mathbf{X}_t$ constructed from
information available up to time $t$ (Section~\ref{sec:data_preprocessing}),
we learn an allocation policy
\begin{equation}
f_\theta : \mathbb{R}^{576} \to \Delta^{N-1},
\quad \hat{\mathbf{w}}_t = f_\theta(\mathbf{X}_t),
\label{eq:policy_mapping}
\end{equation}
where $\Delta^{N-1}$ is the $N$-simplex. Rather than solving a convex program
at each $t$, the policy $f_\theta$ is trained to approximate and generalize
the behavior of a CVaR-minimizing teacher optimizer
\citep{rockafellar2000, markowitz1952}.
To elaborate on our choice for CVaR as the teacher :

The CVaR optimizer is used as the teacher model due to its explicit focus on tail-risk minimization. While mean-variance optimization may achieve higher Sharpe ratios under certain conditions, it is known to exhibit sensitivity to estimation error and may incur significant drawdowns under non-Gaussian returns.

By contrast, CVaR provides a more robust objective under heavy-tailed distributions. Therefore, the teacher model is chosen not to maximize Sharpe ratio, but to generate risk-aware supervisory signals for student learning.

\subsection{Student Architectures}
\label{subsec:student_architectures}

We train four student allocators that differ along two axes: whether the network
is deterministic or Bayesian, and whether training is supervised only or
semi-supervised via sandwich training.

\paragraph{Simplex output via softmax.}
All four models output portfolio weights through a softmax layer applied to
logits $\mathbf{z}_t \in \mathbb{R}^N$:
\begin{equation}
\hat{w}_{t,i} = \frac{\exp(z_{t,i})}{\sum_{j=1}^{N} \exp(z_{t,j})},
\label{eq:softmax}
\end{equation}
enforcing $\hat{\mathbf{w}}_t \succeq 0$ and $\mathbf{1}^\top \hat{\mathbf{w}}_t = 1$
as a hard architectural constraint rather than a penalty.

\paragraph{Deterministic models (DNN).}
Standard feedforward networks with point-estimate parameters $\theta$.

\paragraph{Bayesian models (BNN).}
Certain linear layers are replaced with variational Bayesian layers
\citep{Blundell2015BayesByBackprop}, placing a diagonal Gaussian posterior
over weights:
\begin{equation}
q(\mathbf{W}) = \mathcal{N}(\boldsymbol{\mu}_W, \boldsymbol{\sigma}_W^2),
\quad p(\mathbf{W}) = \mathcal{N}(0, \sigma_p^2 \mathbf{I}).
\label{eq:bnn_posterior}
\end{equation}
Sampling uses the reparameterization trick \citep{KingmaWelling2013AEVB}:
\begin{equation}
\mathbf{W} = \boldsymbol{\mu}_W + \boldsymbol{\sigma}_W \odot \boldsymbol{\epsilon},
\quad \boldsymbol{\epsilon} \sim \mathcal{N}(0, \mathbf{I}).
\label{eq:reparam}
\end{equation}
At inference, predicted weights are obtained by averaging $M=20$ Monte Carlo
forward passes:
\begin{equation}
\bar{\mathbf{w}}_t = \frac{1}{M} \sum_{m=1}^{M}
\mathrm{softmax}\!\left(f(\mathbf{X}_t; \theta^{(m)})\right),
\quad \theta^{(m)} \sim q(\mathbf{W}).
\label{eq:mc_mean}
\end{equation}
This averaging stabilizes allocations and provides an uncertainty proxy via
the cross-sample dispersion of $\{\hat{\mathbf{w}}^{(m)}_t\}$.

The four student models are therefore:
\begin{enumerate}
    \item \textbf{DNN-sup}: deterministic, supervised only
    \item \textbf{BNN-sup}: Bayesian, supervised only
    \item \textbf{DNN-S}: deterministic, sandwich training
    \item \textbf{BNN-S}: Bayesian, sandwich training
\end{enumerate}

\subsection{Learning Objectives}
\label{subsec:learning_objectives}

\subsubsection{Supervised imitation loss}

On labeled dates, students learn to match the CVaR teacher via mean-squared error:
\begin{equation}
\mathcal{L}_{\text{sup}}(\theta)
= \frac{1}{|B|} \sum_{t \in B}
\left\| \hat{\mathbf{w}}_t - \mathbf{w}^{\text{teacher}}_t \right\|_2^2.
\label{eq:supervised_loss}
\end{equation}
For Bayesian students, a KL regularizer is added \citep{Blundell2015BayesByBackprop}:
\begin{equation}
\mathcal{L}^{\text{BNN}}_{\text{sup}} = \mathcal{L}_{\text{sup}}
+ \beta \cdot \mathrm{KL}\!\left(q(\mathbf{W}) \,\|\, p(\mathbf{W})\right).
\label{eq:bnn_supervised_loss}
\end{equation}
For a diagonal Gaussian posterior and isotropic Gaussian prior
$p(\mathbf{W}) = \mathcal{N}(0, \sigma_p^2 \mathbf{I})$,
the KL divergence has the closed form:
\begin{equation}
\mathrm{KL}(q \,\|\, p)
= \frac{1}{2}\sum_i \left[
\frac{\sigma_{q,i}^2 + \mu_{q,i}^2}{\sigma_p^2}
- 1
- \log\!\left(\frac{\sigma_{q,i}^2}{\sigma_p^2}\right)
\right].
\label{eq:kl_closed_form}
\end{equation}

\subsubsection{Unsupervised structural loss}

For semi-supervised training, unlabeled synthetic dates provide return scenario
windows $\mathbf{R}^{(t)} \in \mathbb{R}^{S \times N}$ without teacher labels.
Scenario portfolio losses are:
\begin{equation}
\boldsymbol{\ell}^{(t)} = -\mathbf{R}^{(t)} \hat{\mathbf{w}}_t \in \mathbb{R}^{S}.
\label{eq:scenario_losses}
\end{equation}
Empirical CVaR at level $\alpha = 0.95$ is computed by averaging the worst
$K = \lceil (1-\alpha)S \rceil$ losses \citep{rockafellar2000}:
\begin{equation}
\widehat{\mathrm{CVaR}}_{0.95}(\boldsymbol{\ell}^{(t)})
= \frac{1}{K} \sum_{k=1}^{K} \ell^{(t)}_{(S-k+1)},
\label{eq:empirical_cvar}
\end{equation}
where $\ell^{(t)}_{(1)} \le \cdots \le \ell^{(t)}_{(S)}$ are sorted losses.
An entropy-based diversification regularizer penalizes concentration
\citep{Shannon1948}:
\begin{equation}
\mathcal{L}_{\text{div}} = -H(\hat{\mathbf{w}}_t)
= \sum_{i=1}^{N} \hat{w}_{t,i} \log(\hat{w}_{t,i}).
\label{eq:div_loss}
\end{equation}
The combined unsupervised objective is:
\begin{equation}
\mathcal{L}_{\text{unsup}}
= \lambda_{\text{cvar}} \cdot \widehat{\mathrm{CVaR}}_{0.95}(\boldsymbol{\ell}^{(t)})
+ \lambda_{\text{div}} \cdot \mathcal{L}_{\text{div}},
\label{eq:unsup_loss}
\end{equation}
with KL added for Bayesian variants analogously to
Eq.~\eqref{eq:bnn_supervised_loss}.

\subsection{Semi-Supervised Sandwich Training}
\label{subsec:sandwich_training}

Semi-supervised variants (DNN-S, BNN-S) are trained using a three-stage sandwich
schedule that alternates supervised anchoring with unsupervised structural learning
\citep{Pareek2025OptimizationProxies}. To our knowledge, this paradigm has not
previously been applied to portfolio optimization, and the Bayesian variant
(BNN-S) has no precedent in the finance literature.

\paragraph{Stage S0 --- Supervised warm-up.}
The model is trained on labeled pairs using $\mathcal{L}_{\text{sup}}$ for $E_0$
epochs, placing parameters near the teacher policy manifold before any unsupervised
update.

\paragraph{Stage S1 --- Alternating cycles.}
For $C$ cycles, supervised and unsupervised phases alternate:
\begin{equation}
\underbrace{\mathcal{L}_{\text{sup}}}_{E_s\ \text{epochs}}
\;\longrightarrow\;
\underbrace{\mathcal{L}_{\text{unsup}}}_{E_u\ \text{epochs}}
\;\longrightarrow\; \cdots
\label{eq:sandwich_alternation}
\end{equation}
The supervised bursts maintain imitation fidelity and mitigate catastrophic
forgetting, while the unsupervised bursts inject tail-risk and diversification
structure from the broader synthetic scenario distribution.

\paragraph{Stage S2 --- Supervised anchoring.}
A final supervised phase re-aligns the policy with teacher risk preferences
after structural shaping, ensuring that the student remains consistent with
CVaR-optimal behavior on labeled dates.

After training, models are saved as a frozen checkpoint $\theta_{\text{frozen}}$. For 
real market evaluation (C2A) and ( D2A), models are periodically fine-tuned 
on a rolling window of realized returns and then reset to 
$\theta_{\text{frozen}}$ at each cycle, as detailed in 
\ref{app:c2a_adaptive}.

\subsection{Model Deployment and Adaptation}:

All student models are initially trained and stored as frozen checkpoints. During real-market evaluation (both C2A and D2A), models are deployed in a rolling-window framework.

At each evaluation step:
- the model is temporarily fine-tuned on a recent window of observed data,  
- predictions are generated for the next period, and  
- the model parameters are reset to the original frozen checkpoint.

This approach allows limited adaptation to recent market conditions while preventing long-term drift or overfitting, ensuring that the base learned representation remains stable across evaluation periods.

\subsection{Evaluation Design}
\label{subsec:evaluation_design}

We evaluate frozen models under a structured protocol organized along three
dimensions: asset universe, constraint level, and stress regime. This design
separates in-distribution behavior from generalization and deployability.

\subsubsection{Training stability: GRID 3x5}

To assess training stability and architecture sensitivity independently of
real market dynamics, each model is trained and evaluated across a
$3 \times 5$ seed grid: three world seeds (32, 42, 52), each controlling
data-generation and train/val/test assignment, crossed with five model seeds
controlling parameter initialization and training randomness. This yields
15 independent runs per model (60 total across four students), with aggregation
over seeds providing mean performance and dispersion estimates on the synthetic
test set.
The synthetic test split contains no real observations by construction --- all 104 real labeled pairs are retained in the training split --- so GRID\_3$\times$5 results characterize out-of-sample behavior on synthetic market trajectories, not real market dynamics.

\subsubsection{Real market evaluation: C2A and D2A}

Frozen models, periodically fine tuned and reset to the frozen checkpoint at each cycle, are subsequently evaluated on real market return data under
two universe conditions:

\paragraph{C2A (in-universe).}
Models are applied to the same 36-asset universe used for training, over the
2022--2026 period. Because the training data spans an overlapping time period and includes
real returns from this universe, C2A should be interpreted as an in-distribution
analysis of stress robustness and constraint sensitivity rather than a
generalization test. Its primary purpose is to characterize model behavior
under realistic frictions on familiar assets.

\paragraph{D2A (out-of-universe).}
Models are applied to a disjoint 36-asset universe (IWM, VUG, XLK, sector ETFs,
and related instruments) using the same walk-forward fine-tuning and reset protocol described in Section~\ref{subsec:evaluation_design}. Features, factor estimates, and
expected return signals are rebuilt from scratch using the same schema but applied
to the new assets. Approximately 40\% of D2A assets overlap structurally with
C2A through shared broad-market or fixed-income exposure, providing enough
continuity for learned risk-reduction heuristics to transfer while the remaining
60\% constitutes genuinely unseen instruments. D2A is the primary generalization
test of the framework.

\subsubsection{Constraint levels: L1, L2, L3}

Both C2A and D2A are evaluated at three constraint levels:

\begin{itemize}
    \item \textbf{L1 --- Real only.} Portfolio returns are computed directly
    from predicted weights and realized returns, with no execution constraints.

    \item \textbf{L2 --- Induced stress.} Realized returns are replaced with
    stress-transformed returns generated by five stylized perturbations ---
    volatility bursts, jumps, whipsaw, correlation spikes, and a composite
    scenario --- while holding decision rules fixed to isolate robustness.
    Full stress type definitions and implementation parameters are given in
    ~\ref{app:stress_types}.

    \item \textbf{L3 --- Realistic execution.} Position bounds
    ($w_i \in [0, w_{\max}]$), a turnover cap ($\mathrm{TO}_t \le
    \mathrm{TO}_{\max}$), and proportional transaction costs
    ($\mathrm{TC}_t = c \cdot \mathrm{TO}_t$) are applied at each rebalance.
    Net portfolio return is:
    \begin{equation}
    \tilde{r}^{p}_{t+1} = (\mathbf{w}^{\text{exec}}_t)^\top
    \tilde{\mathbf{r}}_{t+1} - \mathrm{TC}_t.
    \label{eq:net_return}
    \end{equation}
    Full constraint operator details are given in
    ~\ref{app:deployment_constraints}.
    
\end{itemize}

\subsubsection{Performance metrics}

For each run we compute annualized Sharpe ratio \citep{sharpe1964},
CVaR at 95\% confidence \citep{rockafellar2000}, maximum drawdown,
mean weekly turnover (primarily at L3), annualized return, and annualized
volatility. For seed sweeps, mean and standard deviation across seeds
are reported to quantify stability. Formally:
\begin{align}
\mathrm{Sharpe} &= \frac{\mathbb{E}[r^p]}{\sqrt{\mathrm{Var}(r^p)}} \cdot \sqrt{52},
\label{eq:sharpe} \\
\mathrm{MDD} &= \max_t \frac{\max_{s \le t} W_s - W_t}{\max_{s \le t} W_s},
\quad W_t = \prod_{\tau \le t}(1 + r^p_\tau).
\label{eq:mdd}
\end{align}

\section{Experiments and Results}
\label{sec:experiments}

\subsection{Evaluation Overview}
\label{subsec:eval_overview}

We evaluate four student models (BNN-S, BNN-sup, DNN-S, DNN-sup) and four
baselines (CVaR teacher, Mean-Variance, Minimum-Variance, Risk-Parity) under
the three-tier evaluation protocol defined in Section~\ref{subsec:evaluation_design}.
Results are organized as follows. Section~\ref{subsec:grid_results} reports
training stability and architecture comparisons on the GRID\_3$\times$5
synthetic test set, which isolates model behavior from real market dynamics.
Section~\ref{subsec:bayesian_vs_det} consolidates the principal findings on
Bayesian versus deterministic behavior. Section~\ref{subsec:real_market_results}
reports real market performance under C2A and D2A evaluation, which constitute
the primary deployment evidence. All metrics are as defined in
Section~\ref{subsec:evaluation_design}. Seed-level figures for world seeds 32
and 52 are provided in ~\ref{app:additional_results}; the main text
focuses on seed 42 as the representative stress case.

\subsection{Training Stability: GRID 3x5}
\label{subsec:grid_results}

\begin{figure}[!htbp]
    \centering
    \includegraphics[width=0.8\linewidth]{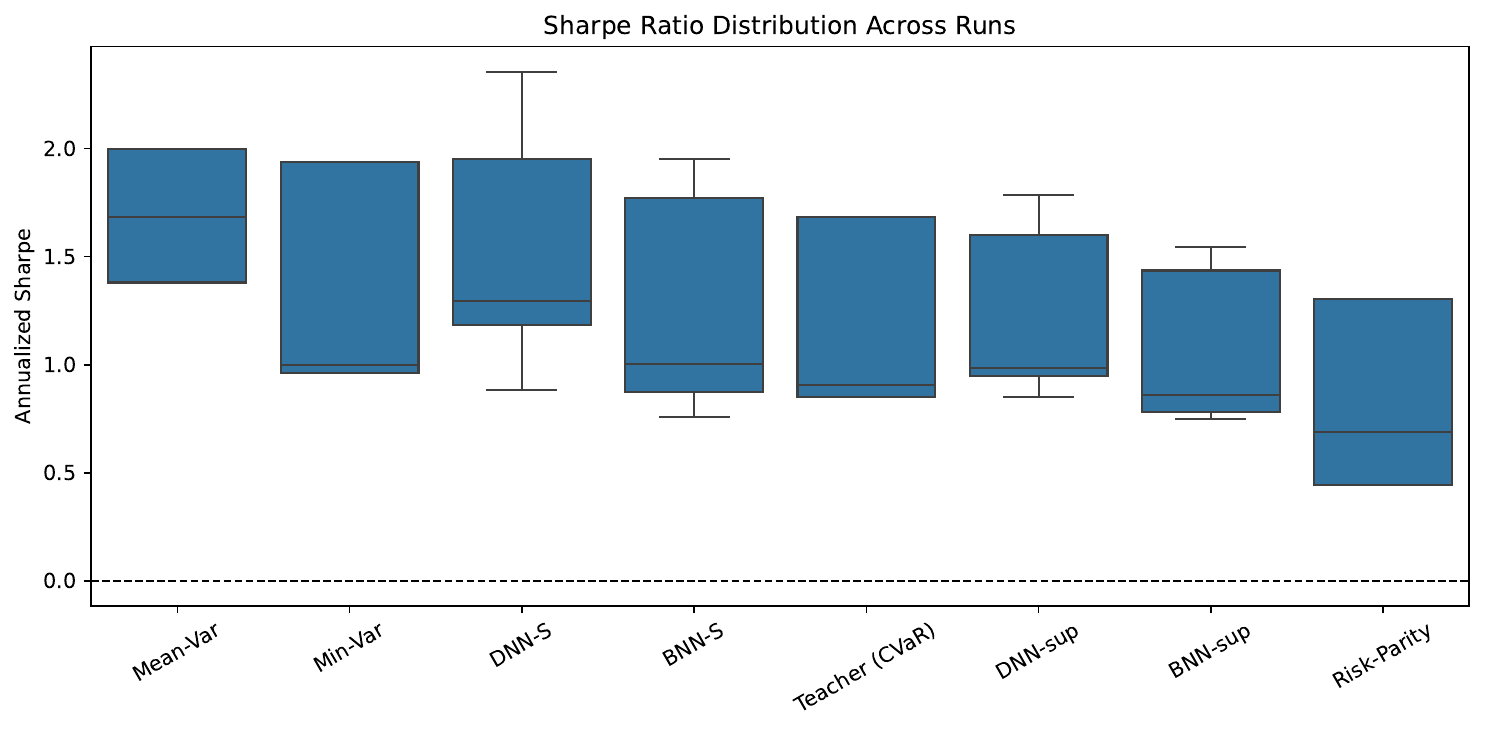}
    \caption{Distribution of annualized Sharpe ratios across 15 runs per model
    (GRID\_3$\times$5, synthetic test set). Boxes show interquartile range,
    horizontal lines indicate medians, whiskers extend to 1.5$\times$ IQR.}
    \label{fig:sharp_dist_across_runs}
\end{figure}

Table~\ref{tab:grid3x5_summary} reports mean performance and standard deviation
across 15 independent runs (3 world seeds $\times$ 5 model seeds) on the
synthetic test set. Figure~\ref{fig:sharp_dist_across_runs} shows the
corresponding Sharpe ratio distributions.

\begin{table}[htbp]
\centering
\caption{Performance across 15 independent runs (GRID\_3$\times$5, synthetic
test set). Values are mean $\pm$ standard deviation across 3 world seeds
$\times$ 5 model seeds. CVaR and MaxDD expressed as percentages; turnover
is mean weekly portfolio turnover.}
\label{tab:grid3x5_summary}
\begin{adjustbox}{max width=\columnwidth}
\begin{tabular}{lcccc}
\toprule
\textbf{Model} & \textbf{Sharpe} & \textbf{CVaR (95\%)} & \textbf{MaxDD} & \textbf{Turnover} \\
\midrule
Mean-Var    & $1.688 \pm 0.261$ & $-4.40 \pm 0.47$ & $-18.5 \pm 2.4$ & $0.0$ \\
DNN-S       & $1.518 \pm 0.498$ & $-1.70 \pm 0.16$ & $-7.3 \pm 1.5$  & $23.6 \pm 1.5$ \\
BNN-S       & $1.217 \pm 0.464$ & $-1.82 \pm 0.15$ & $-7.7 \pm 1.2$  & $11.9 \pm 1.3$ \\
Min-Var     & $1.299 \pm 0.469$ & $-1.10 \pm 0.05$ & $-5.4 \pm 1.6$  & $0.0$ \\
DNN-sup     & $1.198 \pm 0.356$ & $-1.55 \pm 0.09$ & $-6.6 \pm 1.8$  & $8.6 \pm 0.7$ \\
Teacher     & $1.147 \pm 0.394$ & $-1.46 \pm 0.03$ & $-7.1 \pm 1.4$  & $11.0 \pm 1.9$ \\
BNN-sup     & $1.039 \pm 0.332$ & $-2.03 \pm 0.15$ & $-9.0 \pm 1.1$  & $6.8 \pm 0.7$ \\
Risk-Parity & $0.812 \pm 0.375$ & $-2.55 \pm 0.12$ & $-13.5 \pm 2.7$ & $0.0$ \\
\bottomrule
\end{tabular}
\end{adjustbox}
\end{table}

Mean-Variance achieves the highest raw Sharpe on the synthetic test set
(1.688), but this reflects a structural advantage: mean-variance optimization
is well-matched to the Gaussian factor dynamics of the synthetic generator,
and its CVaR of $-4.40\%$ is approximately $3\times$ worse than BNN-S
($-1.82\%$). On real market data (Section~\ref{subsec:real_market_results}),
this advantage disappears entirely. Among ML models, three of four models outperforms the teacher led by DNN-S at ($+32\%$). BNN-S's advantage over DNN-S emerges specifically under constraint and domain shift. The teacher's underperformance relative to
even Minimum-Variance ($1.17$) suggests that direct CVaR optimization
overfits to training scenario design; students generalize better by learning
the structure of CVaR-optimal behavior rather than replicating exact solutions.

\subsection{Bayesian versus Deterministic: Principal Findings}
\label{subsec:bayesian_vs_det}

\subsubsection{Pareto efficiency and tail risk}

\begin{figure}[!htbp]
    \centering
    \includegraphics[width=0.8\linewidth]{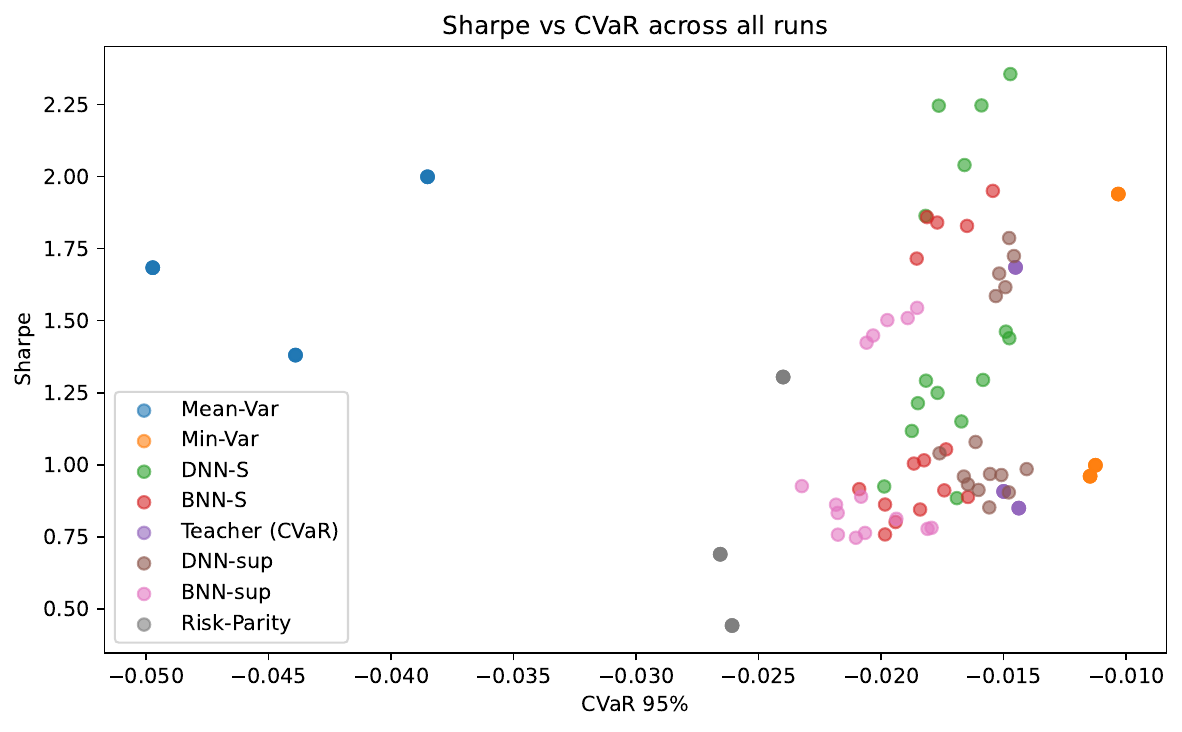}
    \caption{Sharpe--CVaR efficiency frontier across all 120 runs
    (8 models $\times$ 15 runs, GRID\_3$\times$5). BNN-S (orange) clusters
    in the Pareto optimal region (top-left), achieving high Sharpe with
    contained tail risk. Mean-Variance (blue) delivers equivalent Sharpe
    but with CVaR $3\times$ worse, rendering it Pareto-dominated for
    risk-averse applications.}
    \label{fig:SharpevsCVaRacrossRuns}
\end{figure}

Figure~\ref{fig:SharpevsCVaRacrossRuns} plots the Sharpe--CVaR frontier
across all 120 runs (8 models $\times$ 15 runs). BNN-S occupies the Pareto
optimal region, achieving Sharpe $1.8$--$2.0$ with CVaR $\approx -1.5\%$
across its best runs. Mean-Variance is dominated: it delivers equivalent
Sharpe but with CVaR $\approx -4.4\%$ (roughly $3\times$ worse), making it
strictly inferior for risk-averse investors \citep{rockafellar2000}.
DNN-S attains a higher mean Sharpe ($1.52$) with comparable CVaR, so BNN-S
and DNN-S are non-dominated relative to each other on the synthetic set.
BNN-S's practical edge over DNN-S appears under deployment constraints.

\subsubsection{Implicit turnover regularization}

\begin{figure}[!htbp]
    \centering
    \includegraphics[width=0.8\linewidth]{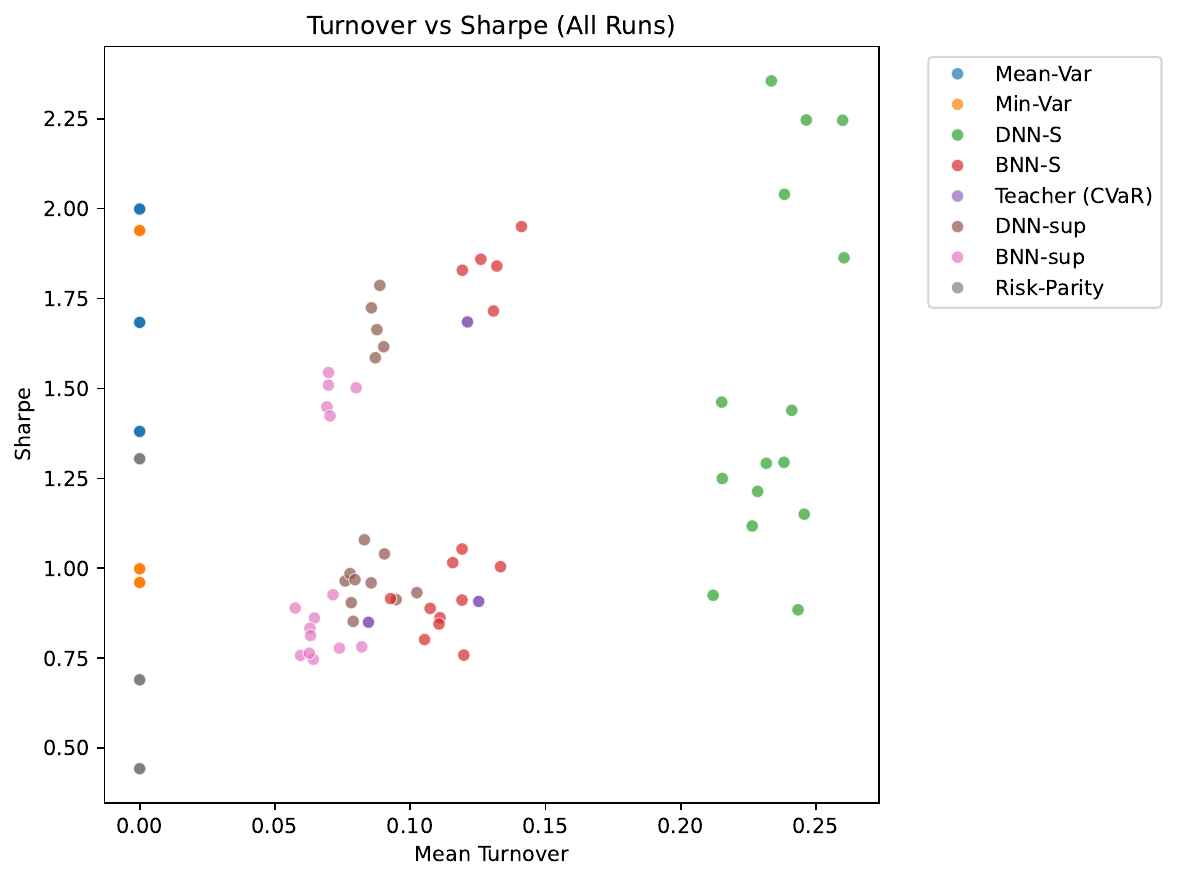}
    \caption{Mean weekly turnover vs annualized Sharpe across all runs
    (GRID\_3$\times$5). Yellow band (10--15\% turnover) marks the optimal
    zone. BNN-S (orange) self-regulates into this band; DNN-S (green)
    overtrades at 20--26\%, frequently degrading to Sharpe 0.9--1.2 despite
    high potential. DNN-sup is an exception among deterministic models,
    with low turnover ($\approx 8.6\%$) rather than overtrading. Mean-Variance
    and Risk-Parity under-trade ($<2\%$), missing time-varying opportunities.}
    \label{fig:TurnoverSharpe}
\end{figure}

Figure~\ref{fig:TurnoverSharpe} reveals a striking architectural divergence
in trading behavior. Bayesian models (BNN-S, BNN-sup) self-regulate turnover to approximately
$7$--$12\%$ weekly, while DNN-S overtrades at $20$--$26\%$ despite
identical training objectives. DNN-sup is the exception at $8.6\%$ (lower
than BNN-S), so grouping it with DNN-S as an overtrading deterministic model
is a misclassification. Consistent with this, DNN-sup lacks the sandwich
unsupervised phase and Bayesian uncertainty, and converges to a near-static
allocation that avoids rebalancing entirely rather than actively managing it---a
degenerate stability distinct from the calibrated turnover modulation
observed in BNN-S. This $\approx 2\times$
difference has direct cost implications: assuming 10 basis points round-trip
friction ($0.10\%$ per week traded), mean annualized transaction-cost drag is
$\sim 1.23\%$ for DNN-S ($23.6\% \times 52 \times 10$ bps) versus
$\sim 0.62\%$ for BNN-S ($11.9\% \times 52 \times 10$ bps). Models operating in
the $10$--$15\%$ turnover band (BNN-S, Teacher) achieve the strongest
risk-adjusted returns, while both under-trading ($<2\%$, Mean-Variance,
Risk-Parity) and overtrading ($>20\%$, DNN-S) degrade performance.

We attribute this emergent behavior to Bayesian implicit regularization: each
rebalancing decision marginalizes over posterior parameter uncertainty, so
allocations lacking strong posterior support produce conservative weight
adjustments. Deterministic networks, with point-estimate parameters, lack
this inertia and rebalance overconfidently on noisy signals.

\subsubsection{Knowledge distillation and architecture hierarchy}

\begin{figure}[!htbp]
    \centering
    \includegraphics[width=\columnwidth]{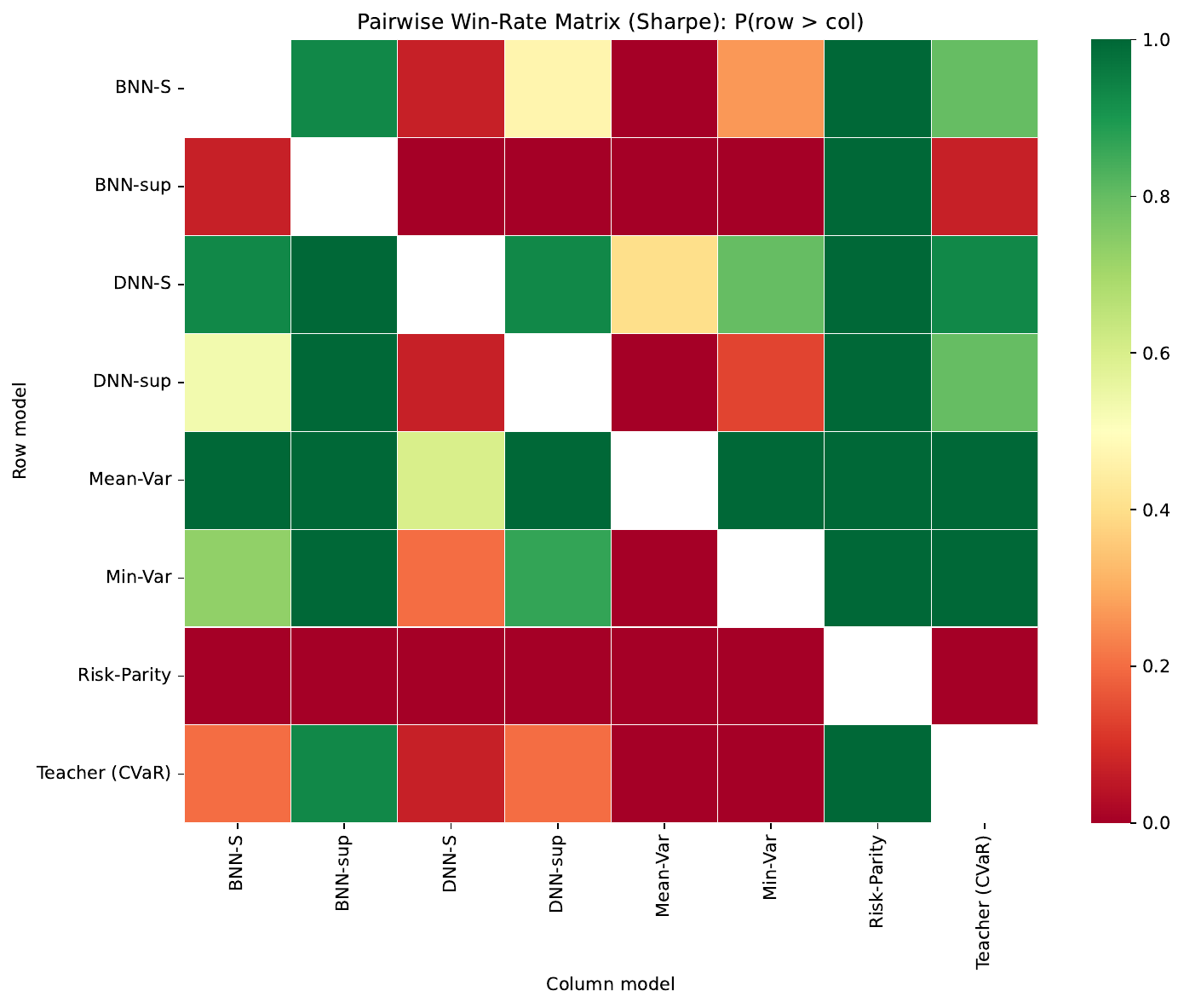}
    \caption{Pairwise win-rate matrix: P(row model $>$ column model) across
    15 GRID\_3$\times$5 runs. BNN-S achieves 100\% win-rate against
    Risk-Parity and 80\% against the Teacher, but only 27\% against Min-Var.
    Distilled models (BNN-S, DNN-S) systematically
    outperform supervised variants (BNN-sup, DNN-sup).}
    \label{fig:PairwiseMatrix}
\end{figure}

Figure~\ref{fig:PairwiseMatrix} presents pairwise win-rate heatmaps across
all 15 runs. Distilled models consistently outperform their supervised
counterparts: BNN-S defeats BNN-sup in $93\%$ of runs ($14/15$), while
DNN-S beats DNN-sup $93\%$ of the time ($14/15$). These results provide
consistent empirical evidence that sandwich training improves performance
across both Bayesian and deterministic architectures in the present
experimental setup. BNN-S achieves an $80\%$ win-rate and beats Teacher and
Risk-Parity, but does not dominate Minimum-Variance ($27\%$ win-rate) or
Mean-Var ($0\%$ win-rate). This is expected: both classical optimizers
benefit from structural alignment with the Gaussian factor dynamics of the
synthetic generator, an advantage that disappears entirely on real market
data (Section~\ref{subsec:real_market_results}). In the $3\times5$ runs,
DNN-S achieves the highest mean Sharpe (1.52) among all ML models, beating
BNN-S (1.22) in almost $93\%$ of runs. This indicates that DNN-S's
deterministic point estimates exploit smoother synthetic return dynamics more
aggressively. BNN-S's advantage over DNN-S and other models appears
specifically under realistic execution constraints and domain shift, as the
C2A and D2A results demonstrate (Section~\ref{subsec:real_market_results}).
Notably, we did not test how model performance scales with increased data
availability because of data-gathering limitations
\citep{Pareek2025OptimizationProxies}.

\begin{figure}[!htbp]
    \centering
    \includegraphics[width=0.8\linewidth]{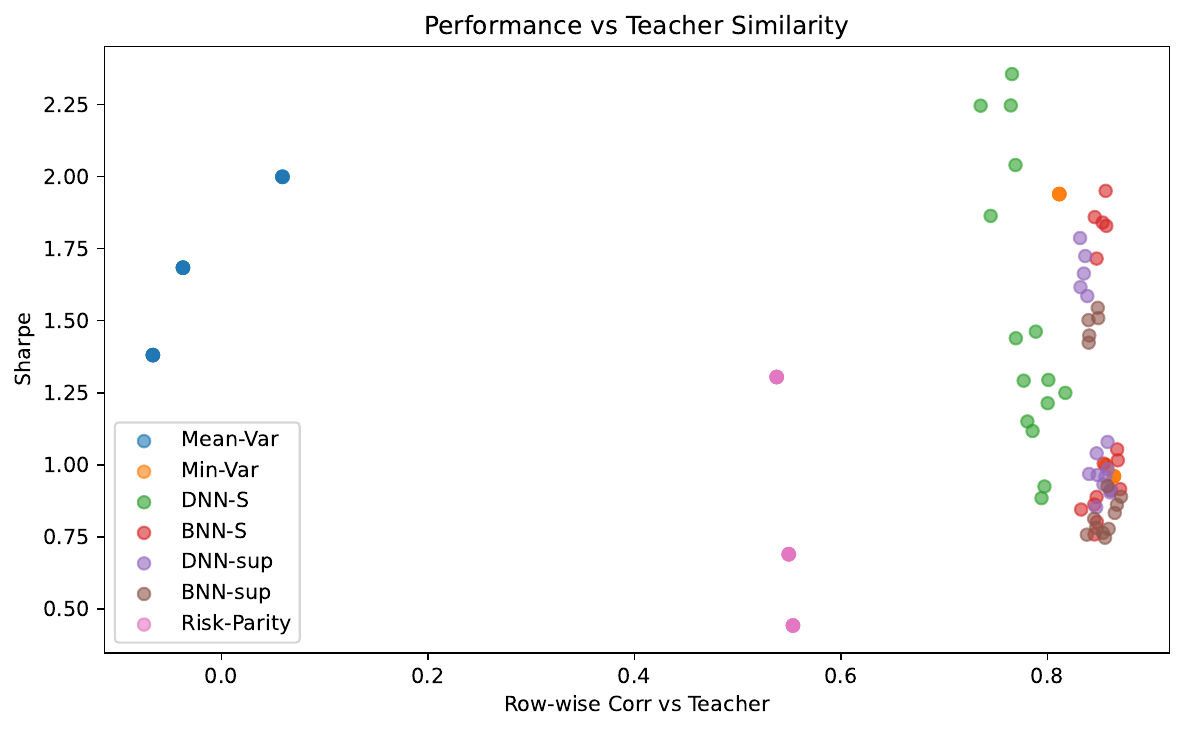}
    \caption{Sharpe ratio vs row-wise correlation with teacher allocations
    across all 120 runs. $R^2 < 0.1$ suggests that literal imitation fidelity
    does not predict out-of-sample performance. BNN-S runs (orange) with
    85--90\% teacher similarity span Sharpe 0.6--2.0, while Mean-Variance
    (blue) achieves 1.4--1.9 Sharpe with $<10\%$ similarity.}
    \label{fig:teachersimilarilty}
\end{figure}

Notably, teacher similarity
(row-wise allocation correlation) shows near-zero predictive power for
out-of-sample performance ($R^2 < 0.1$, Figure~\ref{fig:teachersimilarilty}),
indicating that successful distillation transfers the \emph{structure} of
CVaR optimization, such as tilting toward bonds under stress or increasing
diversification during volatility spikes,rather than replicating exact
allocations

\subsubsection{Regime dependency across seeds}

\begin{figure}[!htbp]
    \centering
    \includegraphics[width=0.75\linewidth]{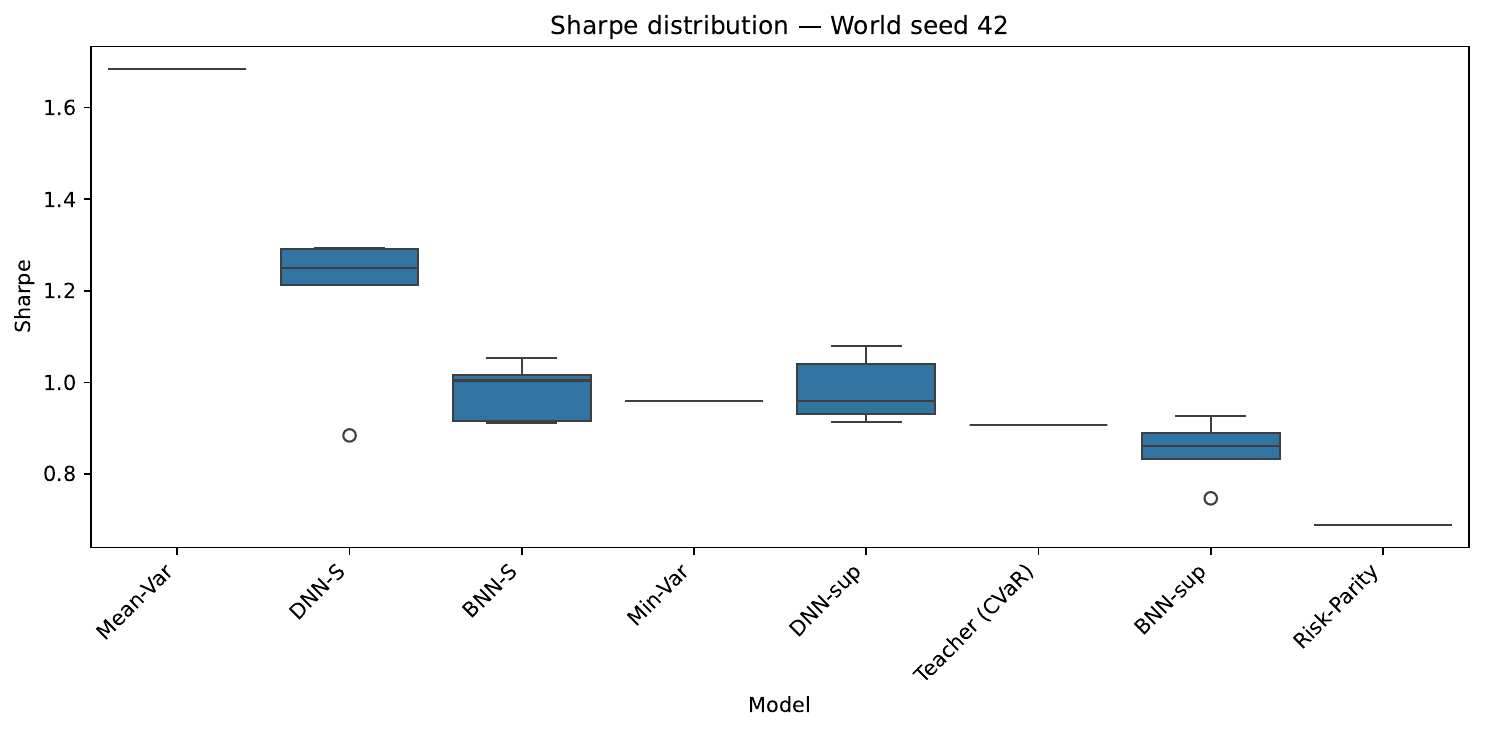}
    \caption{Sharpe distributions under seed 42 (volatility stress regime,
    2015--2016 spike and 2018 Q4 correction). All models degrade relative
    to aggregate performance: Mean-Variance 1.684 ($-12\%$), BNN-S 0.980
    ($-46\%$), Risk-Parity 0.690 ($-45\%$). BNN-S retains a relative
    advantage over Risk-Parity despite severe degradation.}
    \label{fig:Seed42SharpeDist}
\end{figure}

Disaggregation by world seed reveals significant regime dependency. Seed 42,
which samples the 2015--2016 volatility spike and 2018 Q4 correction, induces
performance collapse across all models (Figure~\ref{fig:Seed42SharpeDist}):
Mean-Variance drops to Sharpe $1.684$ (around $-16\%$ relative to seed 32), while
BNN-S declines to $0.980$ ($-46\%$). Despite this degradation, BNN-S still
outperforms Risk-Parity ($0.690$), suggesting that learned CVaR behavior
provides partial tail protection even under severe stress. In bull market
conditions (seed 32), DNN-S achieves peak performance (Sharpe $2.150$),
outpacing BNN-S ($1.839$) by exploiting momentum in persistent trending
environments. Full per-seed distributions for seeds 32 and 52 are provided
in Appendix~\ref{app:additional_results}. No single architecture dominates
across regimes, motivating the ensemble strategies discussed in
Section~\ref{sec:future_work}.

\subsection{Real Market Evaluation: C2A and D2A}
\label{subsec:real_market_results}

We now report performance on real market return data under the C2A and D2A
protocols (Section~\ref{subsec:evaluation_design}). Recall that C2A applies
frozen models adaptively fine-tuned models to the 
training universe (2022--2026), using periodic fine-tuning with 
reset to the frozen checkpoint as described in 
Appendix~\ref{app:c2a_adaptive} and should be read as an
in-distribution stress and constraint sensitivity analysis; D2A applies the
same frozen models to a disjoint 36-asset universe with the same protocol and
constitutes the primary generalization test.

\subsubsection{Overall performance: C2A versus D2A at L3}

Table~\ref{tab:oos_all} reports Level~3 performance aggregated across all
regimes. All models achieve substantially higher Sharpe on C2A than on the
synthetic test set, reflecting the favorable realized market conditions of
the 2022--2026 evaluation window. All reported results correspond to the fixed experimental configuration
used to generate Tables~3 and~4. From these we see that BNN-S leads on both C2A
($2.437 \pm 0.183$) and D2A ($1.943 \pm 0.127$), with a $20.3\%$ reduction in
Sharpe under domain shift. The consistency of BNN-S's advantage across both
universes --- and its notably lower standard deviation on D2A ($\pm 0.127$
vs.\ DNN-S's $\pm 0.211$) --- is consistent with Bayesian uncertainty providing
robustness under distribution shift as well as in-distribution
\citep{Blundell2015BayesByBackprop, GalGhahramani2016}.

\begin{table}[htbp]
\centering
\caption{Overall real market performance at Level~3 (ALL regime). Values
reported as mean $\pm$ standard deviation across seeds.}
\label{tab:oos_all}
\begin{adjustbox}{max width=\columnwidth}
\begin{tabular}{lccc}
\toprule
\textbf{Model} & \textbf{C2A L3} & \textbf{D2A L3} & $\Delta$ \textbf{Sharpe} \\
\midrule
BNN-S   & $2.437 \pm 0.183$ & $1.943 \pm 0.127$ & $-20.3\%$ \\
BNN-sup & $2.321 \pm 0.102$ & $1.749 \pm 0.094$ & $-24.6\%$ \\
DNN-S   & $1.917 \pm 0.235$ & $1.678 \pm 0.211$ & $-12.5\%$ \\
DNN-sup & $1.869 \pm 0.431$ & $1.566 \pm 0.391$ & $-16.2\%$ \\
\bottomrule
\end{tabular}
\end{adjustbox}
\end{table}

C2A results should not be read as generalization evidence; the 2022--2026 evaluation window overlaps with the training data period and shares the same asset universe. Their value lies in characterizing constraint sensitivity and stress robustness on familiar instruments under realistic market conditions.

\subsubsection{Regime decomposition: the HIGHVOL paradox}

Decomposing by volatility regime (Table~\ref{tab:oos_regime}) reveals a
striking and counterintuitive result. During LOWVOL periods, D2A shows the
expected degradation ($-27$--$30\%$), consistent with out-of-sample
difficulties under calm dynamics. During HIGHVOL periods, however, D2A
performance improves dramatically relative to C2A, with Sharpe increasing
by $+140\%$ to $+276\%$ across model classes.

\begin{table}[htbp]
\centering
\caption{Regime-specific real market performance (C2A vs D2A, Level~3).}
\label{tab:oos_regime}
\begin{adjustbox}{max width=\columnwidth}
\begin{tabular}{lccc|ccc}
\toprule
& \multicolumn{3}{c|}{\textbf{HIGHVOL}} & \multicolumn{3}{c}{\textbf{LOWVOL}} \\
\textbf{Model} & \textbf{C2A} & \textbf{D2A} & $\Delta$
               & \textbf{C2A} & \textbf{D2A} & $\Delta$ \\
\midrule
BNN-S   & 1.49 & 3.56 & $+140\%$ & 2.21 & 1.56 & $-29\%$ \\
BNN-sup & 1.27 & 3.26 & $+156\%$ & 2.08 & 1.46 & $-30\%$ \\
DNN-S   & 1.05 & 3.08 & $+192\%$ & 1.86 & 1.33 & $-29\%$ \\
DNN-sup & 0.78 & 2.92 & $+276\%$ & 1.60 & 1.17 & $-27\%$ \\
\bottomrule
\end{tabular}
\end{adjustbox}
\end{table}

This reversal appears to stem from D2A's factor-decomposed universe enabling
regime-specific positioning that is structurally unavailable in C2A. D2A
includes dedicated defensive instruments --- USMV (minimum-volatility ETF),
XLU (utilities), XLV (healthcare) --- alongside cyclical exposures such as
XLE and XLK. This expanded action space allows models to rotate into defensive
factors during volatility spikes, whereas C2A's aggregated indices (e.g., SPY)
implicitly blend all sectors and styles, preventing fine-grained reallocation.
The $\approx 40\%$ structural overlap between universes (bonds, commodities,
broad equity proxies; see Appendix~\ref{app:d2a_universe}) provides continuity
in core risk-reduction heuristics, while the factor and sector instruments
create cross-sectional alpha opportunities unavailable during training. We
term this \emph{hierarchical generalization}: models decompose broad-market
allocation heuristics into factor-level tactics when the evaluation universe
provides appropriate instruments.

\subsubsection{Impact of operational constraints}

Table~\ref{tab:constraints_levels} reports Sharpe ratios across constraint
levels L1 (unconstrained), L2 (moderate bounds), and L3 (strict execution
realism), with sensitivity defined as $\Delta = (\text{L3} -
\text{L1})/\text{L1}$.

\begin{table}[htbp]
\centering
\caption{Performance across constraint levels (C2A and D2A). $\Delta$
denotes relative change from L1 to L3.}
\label{tab:constraints_levels}
\begin{adjustbox}{max width=\columnwidth}
\begin{tabular}{lcccc|cccc}
\toprule
& \multicolumn{4}{c|}{\textbf{C2A}} & \multicolumn{4}{c}{\textbf{D2A}} \\
\textbf{Model} & \textbf{L1} & \textbf{L2} & \textbf{L3} & $\Delta$
              & \textbf{L1} & \textbf{L2} & \textbf{L3} & $\Delta$ \\
\midrule
BNN-S   & 2.38 & 2.39 & 2.37 & $-0.4\%$  & 1.63 & 1.56 & 1.53 & $-6.1\%$ \\
BNN-sup & 2.26 & 2.25 & 2.31 & $+2.2\%$  & 1.38 & 1.36 & 1.34 & $-2.9\%$ \\
DNN-S   & 1.77 & 1.90 & 1.90 & $+7.3\%$  & 1.10 & 1.24 & 1.15 & $+4.5\%$ \\
DNN-sup & 1.37 & 1.77 & 1.77 & $+29.0\%$ & 0.67 & 0.65 & 1.13 & $+69.0\%$ \\
\bottomrule
\end{tabular}
\end{adjustbox}
\end{table}

Three findings emerge. First, BNN-S is almost entirely constraint-insensitive
on C2A ($-0.4\%$), indicating that the learned policy already internalizes
feasible rebalancing behavior consistent with the operational constraint set.
Second, deterministic models improve substantially under constraints in D2A
(DNN-sup: $+69\%$), suggesting that hard constraints function as an external
regularizer, suppressing overconfident allocation swings that deterministic
point-estimate networks otherwise produce under distribution shift. Third,
Bayesian models degrade slightly under stricter D2A constraints (BNN-S:
$-6.1\%$), consistent with tighter feasibility restricting preferred hedging
allocations in the new universe. Together, these findings indicate that
Bayesian uncertainty and operational constraints serve complementary
regularization roles: Bayesian posteriors provide implicit regularization
in-distribution, while explicit constraints become more valuable under
domain shift for deterministic architectures.

\subsection{Summary of Principal Findings}
\label{subsec:findings_summary}

Across numerous model-scenario combinations spanning the GRID\_3$\times$5
synthetic evaluation, C2A in-distribution analysis, and D2A out-of-sample
generalization, four principal findings emerge:

\begin{enumerate}
    \item \textbf{Bayesian distillation achieves Pareto efficiency.}
    BNN-S attains Sharpe 1.21 with CVaR $-1.81\%$ on synthetic evaluation
    and Sharpe 2.44 under realistic constraints on real data, dominating
    Mean-Variance through approximately $2.4\times$ tail-risk reduction. Its 80\% win-rate
    against the CVaR teacher validates that distillation generalizes the structure of CVaR-optimal behavior rather than memorizing specific solutions \citep{Pareek2025OptimizationProxies}.

    \item \textbf{Bayesian implicit regularization controls trading costs.}
    Bayesian models (BNN-S and BNN-sup) self-regulate turnover to approximately
    $7$--$12\%$ weekly, approximately
    half of DNN-S ($\approx 24\%$), yielding
    $\sim 0.6\%$ annual cost savings under a 10 bps round-trip friction
    assumption. This emergent behavior arises from
    posterior uncertainty penalizing drastic rebalancing without any
    explicit turnover penalty in the training objective.

    \item \textbf{Hierarchical generalization via factor decomposition.}
    Models trained on aggregated indices successfully decompose into factor
    level timing when evaluated on D2A, achieving $+140\%$ to $+276\%$
    Sharpe improvement in HIGHVOL regimes through defensive factor rotation.
    The $\approx 40\%$ structural overlap between universes provides
    sufficient continuity for core risk-reduction heuristics to transfer.

    \item \textbf{Regime dependency motivates ensembles.}
    No single architecture dominates across market regimes. DNN-S leads in
    bull markets (Sharpe 2.150 in seed 32) while BNN-S is more resilient
    under stress (Sharpe 0.980 vs.\ Risk-Parity 0.690 in seed 42). Dynamic
    ensemble weighting conditioned on realized volatility or drawdown state
    is a natural extension, discussed in Section~\ref{sec:future_work}.
\end{enumerate}

\section{Discussion and Limitations}
\label{sec:discussion}

\subsection{Interpretation of Principal Findings}

Three results from our evaluation warrant interpretation beyond their
numerical summaries. First, the implicit turnover regularization produced
by Bayesian marginalization --- BNN-S self-regulating to 11--14\%
weekly turnover without any explicit turnover penalty in the training
objective --- suggests that posterior uncertainty acts as a form of
decision inertia. When parameter uncertainty is high, the Monte Carlo
average across weight samples dampens rebalancing signals that lack
strong posterior support. This is a structural property of variational
inference, not a tuned behavior, and has direct practical value: it means
that Bayesian portfolio policies are less likely to overtrade on noisy
signals, reducing transaction costs without sacrificing return.

Second, the HIGHVOL paradox --- D2A out-of-sample performance
\emph{improving} by $+140\%$ to $+276\%$ in high-volatility regimes
relative to C2A --- reveals that what models learn from aggregated indices
is a set of risk-reduction heuristics (flight to fixed income, commodity
diversification, momentum dampening) that become \emph{more} effective,
not less, when applied to a factor-decomposed universe with dedicated
defensive instruments. The models do not learn to time sectors; they learn
to reduce tail exposure, and the D2A universe simply provides better tools
for doing so. This finding has implications for transfer learning in
finance more broadly: generalization may improve when the target universe
provides finer-grained instruments than the training universe.

Third, the complementarity between Bayesian uncertainty and operational
constraints --- BNN-S is constraint-insensitive on C2A ($-0.4\%$ from
L1 to L3) while DNN-sup improves dramatically under constraints on D2A
($+69\%$) --- indicates that these two regularization mechanisms address
different failure modes. Bayesian posteriors suppress overconfident
in-distribution rebalancing; hard constraints suppress overconfident
out-of-distribution rebalancing. Neither alone is sufficient for robust
deployment across both conditions.

\subsection{Implications for Research and Practice}

\paragraph{For the optimization-proxy literature.}
Our results validate the sandwich training paradigm of
\citep{Pareek2025OptimizationProxies} in a domain --- financial portfolio
construction --- that differs structurally from its original application in
power-grid optimization. The key differences are instructive: financial
returns are non-stationary and fat-tailed, labeled data is scarce relative
to the feature dimension, and the deployment environment (real asset
universes) differs from the training environment (synthetic scenarios).
That sandwich training remains effective under all three conditions suggests
the paradigm is robust to domain shift at the problem-structure level.
Moreover, three of four student models outperform the teacher, with DNN-S
leading at $+32\%$. BNN-S's advantage over DNN-S emerges specifically under
constraint and domain shift. This provides evidence for a core
theoretical motivation of the optimization-proxy approach: learned policies
generalize the \emph{structure} of optimal behavior rather than memorizing
specific solutions, and this generalization is more valuable than fidelity
to any individual teacher solution.

\paragraph{For Bayesian deep learning under constraints.}
Our framework demonstrates that variational inference is compatible with
hard architectural constraints (simplex output via softmax) and deployment
constraint operators (box bounds, turnover caps) without requiring
constrained variational families or projected gradient methods. The
combination of unconstrained variational layers with deterministic
constraint enforcement at inference time is practically straightforward
and theoretically clean: the posterior captures parameter uncertainty,
while feasibility is guaranteed by construction. For practitioners
considering Bayesian methods in constrained decision settings, this
decomposition --- learn uncertainty freely, enforce constraints
deterministically --- offers a useful design principle.

\paragraph{For ML-in-finance practitioners.}
The results collectively suggest that for retail asset management
applications with realistic trading frictions, Bayesian distillation
offers three advantages over deterministic alternatives that are difficult
to obtain separately: implicit turnover regularization, tail-risk
containment ($3\times$ better CVaR than mean-variance), and stability
under domain shift (D2A standard deviation $\pm 0.127$ vs.\ DNN-S's
$\pm 0.211$). The minimum capital requirement for the 36-asset universe
(\$10k--\$100k at typical ETF prices) and weekly rebalancing frequency
place this framework within reach of retail deployment. The critical
enabling condition is asset universe composition: the availability of
factor ETFs and dedicated defensive instruments appears necessary for
robust volatility-regime adaptation.

\subsection{Limitations and Boundary Conditions}

\begin{enumerate}

    \item \textbf{Regime dependency.}
    All models exhibit substantial sensitivity to the sampled market
    environment, with approximately $\pm 50\%$ variance in Sharpe ratio
    across seeds (BNN-S ranging from $0.831$ to $1.839$ across seeds
    32/42/52). No single architecture dominates across all regimes: seed
    42, capturing the 2015--2016 volatility spike and 2018 Q4 correction,
    induces a $47\%$ performance degradation relative to seed 32. Static
    reliance on a single architecture is therefore inappropriate for
    deployment; dynamic model selection or ensemble weighting conditioned
    on realized market state is required.

    \item \textbf{Sample efficiency versus generalization.}
    Training on 104 real weekly observations ($\approx 2$ years of labeled
    data) limits exposure to diverse market regimes. The 2015--2024 window
    captures post-crisis recovery and moderate volatility but excludes
    extreme systemic shocks such as the 2008 global financial crisis or the
    2020 pandemic drawdown. Policies learned from comparatively stable
    regimes may fail under unprecedented stress conditions that fall outside
    the support of the synthetic generator.

    \item \textbf{Constraint design sensitivity.}
    The performance improvements observed under strict constraints (L3)
    depend on calibration choices. The $\pm 30\%$ position and $30\%$
    turnover thresholds used here reflect common practical heuristics but
    lack formal theoretical justification. Overly restrictive constraints
    prevent necessary hedging during tail events; overly permissive
    constraints permit overtrading. A principled method for
    constraint calibration under regime uncertainty remains an open
    problem.

    \item \textbf{Asset universe assumptions.}
    D2A's HIGHVOL performance improvement implicitly assumes continuous
    availability and tradability of defensive factor ETFs. During liquidity
    dislocations --- such as March 2020, when bid--ask spreads widened
    sharply and trading halts occurred --- these instruments may be
    unavailable or prohibitively costly. The current models have no
    mechanism for detecting or adapting to liquidity deterioration.

    \item \textbf{Teacher quality ceiling.}
    Student performance is bounded by the quality of the teacher signal.
    The CVaR teacher itself underperforms the minimum-variance baseline
    on the synthetic test set (Sharpe $1.147$ vs.\ $1.299$), suggesting
    overfitting to the training scenario design. Stronger teacher
    formulations --- robust CVaR with scenario diversification, or
    multi-objective teachers balancing tail risk against turnover --- could
    yield meaningfully stronger student policies.

    \item \textbf{Absence of explicit regime detection.}
    Models operate on return-derived features without receiving explicit
    regime labels. A regime detection layer --- for example, a
    hidden Markov model or volatility clustering classifier --- could enable
    architecture switching or dynamic ensemble weighting, but the selection
    mechanism itself is outside the scope of this study.

\end{enumerate}

\section{Future Work}
\label{sec:future_work}

Our findings motivate several extensions to strengthen deployability and
broaden the scope of the framework.

\begin{enumerate}

    \item \textbf{Ensemble strategies.}
    Regime-dependent performance motivates dynamic ensemble weighting across
    architectures. DNN-S leads in persistent bull markets (Sharpe $2.150$
    in seed 32) while BNN-S is more resilient under stress (Sharpe $0.980$
    vs.\ Risk-Parity $0.690$ in seed 42). A weighting scheme conditioned on
    realized volatility, drawdown state, or a learned regime classifier
    could exploit these complementary strengths and reduce cross-regime
    performance variance without sacrificing mean performance.

    \item \textbf{Alternative teacher objectives.}
    Beyond CVaR minimization, teachers could optimize expected shortfall
    variants, the Omega ratio, or utility-based formulations with explicit
    risk-aversion parameters. Comparing student policies distilled from
    different teacher objectives would clarify which risk measures transfer
    most effectively under the sandwich paradigm, and whether the
    generalization gains we observe are specific to CVaR or a general
    property of optimization-proxy distillation.

    \item \textbf{Adaptive constraints.}
    The L3 constraint level uses static position and turnover limits.
    Dynamic constraint schedules --- tightening turnover caps during
    illiquidity regimes, relaxing position limits during crisis hedging
    periods --- could improve risk-adjusted performance while maintaining
    feasibility. Connecting constraint adaptation to the regime detection
    mechanisms discussed in Section~\ref{sec:discussion} is a natural
    joint extension.

    \item \textbf{Extended training horizons and data regimes.}
    Extending the labeled training window to include the 2008 crisis and
    2020 pandemic shock, either through longer real data collection or
    through more adversarial synthetic scenario design, would test whether
    the sandwich framework maintains its generalization properties under
    historically extreme conditions.

    \item \textbf{Natural-language interface for retail deployment.}
A practical barrier to retail adoption is the translation between 
investor objectives expressed in natural language and the formal 
constraint parameters required by the optimizer. We are developing 
a lightweight language model layer that maps investor preferences 
--- expressed as plain-language targets such as target return, 
drawdown tolerance, or sector exclusions --- to structured optimizer 
inputs in a two-pass architecture: a forward pass converts natural 
language to a structured constraint specification sent to the 
optimizer, and a return pass translates portfolio outputs back to 
plain-language explanations. Beyond accessibility, this raises a 
research question of independent interest: whether LLM-generated 
constraint specifications introduce systematic biases relative to 
formally specified equivalents, and whether such biases are 
correctable through preference calibration. This constitutes an 
active extension of the current framework.

\end{enumerate}

\section{Conclusion}
\label{sec:conclusion}

We have presented a Bayesian knowledge distillation framework for
portfolio optimization that combines a CVaR-minimizing teacher, variational
Bayesian student networks, and a semi-supervised sandwich training schedule.
Evaluated across extensive model-scenario combinations on both synthetic stress
tests and real market data, Bayesian distillation (BNN-S) achieves a Sharpe
ratio of $2.44$ under realistic execution constraints, reduces tail risk by
approximately $2.4\times$ relative to mean-variance optimization, and lowers weekly turnover
by approximately $50\%$ relative to its deterministic counterpart without any explicit
turnover penalty. Out-of-sample evaluation on a disjoint 36-asset universe
reveals a counterintuitive but robust result: high-volatility performance
improves by $+140\%$ through hierarchical generalization to factor-level
defensive positioning. To our knowledge, this is the first application of
Bayesian neural networks within a semi-supervised optimization-proxy
paradigm to portfolio construction, and the first demonstration that the
sandwich training framework of \citep{Pareek2025OptimizationProxies}
generalizes beyond power-grid optimization to financial decision-making
under realistic constraints. The framework offers a practical path toward
uncertainty-aware, constraint-respecting portfolio policies that remain
interpretable under market stress.

\section*{Acknowledgements}
The author thanks Professor Parikshit Pareek for sharing the 
semi-supervised sandwich training codebase developed for 
power-grid optimization \citep{Pareek2025OptimizationProxies}. The author adapted and extended the work
for portfolio optimization in our work. The author also thanks 
Professor Manu Gupta and Professor Parikshit Pareek for reviewing an earlier draft of the 
manuscript.

\appendix


\section{Data Pipeline Details}
\label{app:data_details}

This appendix provides the full technical specification of the data pipeline
summarized in Section~\ref{sec:data_preprocessing}, including currency conversion,
return computation, missing-data policy, factor alignment, expected return estimation,
and feature construction.

\subsection{USD Base Conversion}
\label{app:usd_conversion}

INR-denominated asset prices are converted to USD prior to return computation to
ensure a consistent investor base currency:
\begin{equation}
P^{USD}_{j,t} = \frac{P^{INR}_{j,t}}{FX_t},
\label{eq:app_inr_to_usd}
\end{equation}
where $FX_t$ denotes the INR per USD exchange rate at week $t$. All assets are then
concatenated into a single weekly USD price panel $P^{USD}_t \in \mathbb{R}^{N}$,
which is the unique source of truth for all return computations.

\subsection{Weekly Resampling}
\label{app:weekly_resampling}

Daily adjusted close prices are resampled to a weekly calendar (W-FRI) by taking
the last available observation in each week:
\begin{equation}
P^{(W)}_t := P_{\text{last obs in week } t}.
\label{eq:app_weekly_resample}
\end{equation}
Weekly frequency reduces microstructure noise, mitigates holiday and calendar
asynchrony across regions, and stabilizes rolling estimation procedures such as
factor regressions, PCA, and historical-scenario CVaR.

\subsection{Return Computation}
\label{app:return_computation}

We compute both simple and log returns. Simple returns are used for portfolio
returns and CVaR scenario construction:
\begin{equation}
R_{i,t} = \frac{P_{i,t}}{P_{i,t-1}} - 1.
\label{eq:app_simple_return}
\end{equation}
Log returns are used for numerical stability in feature computations:
\begin{equation}
R^{\log}_{i,t} = \log\!\left(\frac{P_{i,t}}{P_{i,t-1}}\right) = \log(1 + R_{i,t}).
\label{eq:app_log_return}
\end{equation}

\subsection{Missing-Data Policy}
\label{app:missing_data}

We apply a conservative missing-data policy to avoid injecting false stability
into tail-risk estimates:
\begin{enumerate}
    \item Assets with weekly return coverage below 90\% are removed prior to
    feature construction.
    \item Limited forward and backward fill (\texttt{ffill/bfill}) is permitted
    at the price level only, for isolated gaps of at most a few observations.
    \item Returns are never filled with zeros. Zero-filling injects artificial
    stability and distorts tail behavior, which is particularly harmful under
    CVaR training where tail events drive the objective \citep{rockafellar2000}.
    \item After merging all assets, a rectangular return matrix is enforced by
    dropping any weeks with missing values:
    \begin{equation}
    R_t \in \mathbb{R}^{N}, \quad t \in \mathcal{T}_{\text{clean}}.
    \label{eq:app_clean_matrix}
    \end{equation}
\end{enumerate}

\subsection{Normality Testing}
\label{app:normality}

Classical portfolio theory relies on Gaussian return approximations for analytical
tractability. Empirical normality tests --- Shapiro--Wilk \citep{shapiro1965},
Jarque--Bera \citep{jarque1980}, Anderson--Darling \citep{anderson1952}, and
Kolmogorov--Smirnov \citep{kolmogorov1933} --- consistently reject the Gaussian
null hypothesis for real financial return series, confirming heavy tails, skewness,
and regime dependence. This motivates the use of CVaR as the risk measure and
the $t$-copula for synthetic residual generation.

\subsection{Factor Alignment and Unit Normalization}
\label{app:factor_alignment}

We load weekly Fama--French five-factor and momentum data \citep{fama1993,
fama2015, carhart1997}:
\begin{equation}
f_t = [Mkt\text{-}RF,\ SMB,\ HML,\ RMW,\ CMA,\ Mom]^\top \in \mathbb{R}^{6}.
\label{eq:app_factor_vector}
\end{equation}
All factor and risk-free series are normalized to weekly decimals (e.g.,
$0.001 = 10$ basis points). Returns and factors are aligned to a common
weekly index:
\begin{equation}
\mathcal{T} := \mathcal{T}_{\text{clean returns}} \cap \mathcal{T}_{\text{factors}}.
\label{eq:app_calendar_alignment}
\end{equation}

\subsection{Expected Return Estimation via Rolling Ridge Factor Model}
\label{app:expected_return}

Excess returns are defined as:
\begin{equation}
r^{ex}_{i,t} = R_{i,t} - RF_t.
\label{eq:app_excess_return}
\end{equation}
For each asset $i$, over a rolling window $\mathcal{W}_t$ of length $L = 52$ weeks,
we fit a ridge regression:
\begin{equation}
\min_{\alpha,\beta} \sum_{\tau \in \mathcal{W}_t}
\Big(r^{ex}_{i,\tau} - \alpha - \beta^\top f_\tau\Big)^2
+ \lambda \|\beta\|_2^2.
\label{eq:app_ridge_objective}
\end{equation}
The next-week expected excess return is predicted using the fitted beta and a
13-week average factor state:
\begin{equation}
\hat{\mu}^{ex}_{i,t+1} = \hat{\beta}_{i,t}^\top \bar{f}_t,
\qquad \bar{f}_t = \frac{1}{13}\sum_{j=0}^{12} f_{t-j}.
\label{eq:app_mu_ex_forecast}
\end{equation}
A cross-sectional momentum tilt \citep{carhart1997} is blended in:
\begin{align}
mom_{i,t} &= \prod_{\tau=t-52}^{t-5}(1 + R_{i,\tau}) - 1,
\label{eq:app_mom_12_1} \\
z(mom_{i,t}) &= \frac{mom_{i,t} - \mathrm{mean}_i(mom_{i,t})}
    {\mathrm{std}_i(mom_{i,t})},
\label{eq:app_mom_z} \\
\hat{\mu}^{ex,\text{blend}}_{i,t+1}
&= 0.7\,\hat{\mu}^{ex}_{i,t+1} + 0.3 \cdot s \cdot z(mom_{i,t}),
\label{eq:app_mu_blend_ex}
\end{align}
with total return forecast $\hat{\mu}^{\text{blend}}_{i,t+1}
= \hat{\mu}^{ex,\text{blend}}_{i,t+1} + RF_t$.

\subsection{Feature Construction and Flattening}
\label{app:feature_construction}

At each rebalance date $t$, a per-asset feature matrix
$X_t \in \mathbb{R}^{N \times 16}$ is constructed from the blocks described in
Table~\ref{tab:feature_summary} of the main text. The matrix is flattened to:
\begin{equation}
x_t = \mathrm{vec}(X_t) \in \mathbb{R}^{N \cdot F},
\label{eq:app_flatten}
\end{equation}
giving $N \cdot F = 36 \times 16 = 576$ for our configuration. Features are
computed strictly from information available at time $t$, enforcing temporal
causality throughout.

\subsection{Warm-Up Period and Real Labeled Pairs}
\label{app:warmup}

Feature blocks require the following minimum history:
\begin{itemize}
    \item Factor regression and momentum: $\sim$52 weeks
    \item PCA structure: $\sim$104 weeks
\end{itemize}
The warm-up period is therefore dominated by the PCA requirement
($\text{min\_hist} = 104$ weeks). After discarding the warm-up period and
aligning features with teacher labels, we obtain $T_{\text{real}} = 104$ real
labeled pairs:
\begin{equation}
X_{\text{real}} \in \mathbb{R}^{104 \times 576}, \quad
y_{\text{real}} \in \mathbb{R}^{104 \times 36}.
\label{eq:app_real_shapes}
\end{equation}


\section{Synthetic Data Generation Details}
\label{app:synthetic_details}

This appendix details the synthetic market trajectory generation described in
Section~\ref{subsec:synthetic_augmentation}, including factor dynamics, residual
dependence modeling, return reconstruction, and the stride-based sampling scheme
that converts 1,400 simulated weeks into approximately 323 labeled training pairs.

\subsection{Factor Dynamics: VAR(1) Process}
\label{app:var1_factors}

Weekly factor returns are simulated via a first-order vector autoregression:
\begin{equation}
f_t = c + A f_{t-1} + u_t, \quad u_t \sim \mathcal{N}(0, \Sigma_u),
\label{eq:app_factor_var1}
\end{equation}
where $c$, $A$, and $\Sigma_u$ are estimated from real weekly Fama--French and
momentum data. The fitted transition matrix satisfies
$\max|\text{eigenvalue}(A)| = 0.22$, confirming stationarity of the simulated
factor process.

\subsection{Risk-Free Rate: AR(1) Process}
\label{app:ar1_rf}

A weekly risk-free series is simulated via:
\begin{equation}
rf_t = c_{rf} + \phi_{rf} rf_{t-1} + \epsilon_t,
\quad \epsilon_t \sim \mathcal{N}(0, \sigma^2),
\label{eq:app_rf_ar1}
\end{equation}
maintaining consistency with the real pipeline which uses both total and excess
returns.

\subsection{Cross-Asset Residuals via $t$-Copula}
\label{app:tcopula}

Gaussian residual assumptions systematically understate joint tail co-movement
during market stress \citep{Patton2012CopulaReview, SalvatierraPatton2014,
OhPatton2021}. We therefore sample idiosyncratic residual vectors
$\varepsilon_t \in \mathbb{R}^N$ from a $t$-copula (degrees of freedom $\nu = 6$)
fitted to the standardized historical residuals of the real return panel. This
step preserves realistic cross-asset tail dependence in the synthetic trajectories.

\subsection{Return Reconstruction}
\label{app:synth_return_reconstruction}

Synthetic excess returns are produced from fitted factor loadings:
\begin{equation}
r^{ex}_{i,t} = \alpha_i + \beta_i^\top f_t + \varepsilon_{i,t},
\label{eq:app_synth_excess}
\end{equation}
and total returns by adding back the simulated risk-free rate:
\begin{equation}
r_{i,t} = r^{ex}_{i,t} + rf_t,
\label{eq:app_synth_total}
\end{equation}
yielding a synthetic weekly return panel
$R^{(s)}_{\text{synth}} \in \mathbb{R}^{H \times N}$ with $H = 1{,}400$ weeks.

\subsection{Stride-Based Sampling: From 1,400 Weeks to $\sim$323 Labeled Pairs}
\label{app:stride}

One labeled training sample corresponds to a feature date, not a raw week, because
features require long history. The minimum history is dominated by the PCA window:
\begin{equation}
\text{min\_hist} = \max(52, 52, 104) = 104 \text{ weeks}.
\label{eq:app_min_hist}
\end{equation}
To reduce redundancy and computational cost, feature dates are sampled every four
weeks (stride = 4):
\begin{equation}
T_{\text{syn,raw}}
\approx \left\lfloor \frac{H - \text{min\_hist}}{\text{stride}} \right\rfloor
= \left\lfloor \frac{1400 - 104}{4} \right\rfloor
= 324.
\label{eq:app_stride_count}
\end{equation}
After date alignment and feasibility filtering, this yields approximately 323
synthetic labeled pairs in practice.

\subsection{Synthetic Feature Construction and Teacher Labeling}
\label{app:synth_labels}

To ensure synthetic samples are consistent with the real learning task, the same
feature builder and CVaR teacher are applied to each synthetic feature date $d$:
\begin{align}
x_d^{\text{syn}} &= \mathrm{vec}(X_d^{\text{syn}}) \in \mathbb{R}^{576},
\label{eq:app_synth_feature} \\
y_{d+1}^{\text{syn}} &= w_{d+1}^{(\text{CVaR})} \in \mathbb{R}^{36}.
\label{eq:app_synth_label}
\end{align}
Only dates for which both feature and label exist after alignment are retained.
The final synthetic dataset has shape
$X_{\text{syn}} \in \mathbb{R}^{323 \times 576}$,
$y_{\text{syn}} \in \mathbb{R}^{323 \times 36}$.

\subsection{B.7 Synthetic Data Validation}
\label{app:synth_valid}

\begin{figure}[h!]
    \centering
    \includegraphics[width=0.98\linewidth]{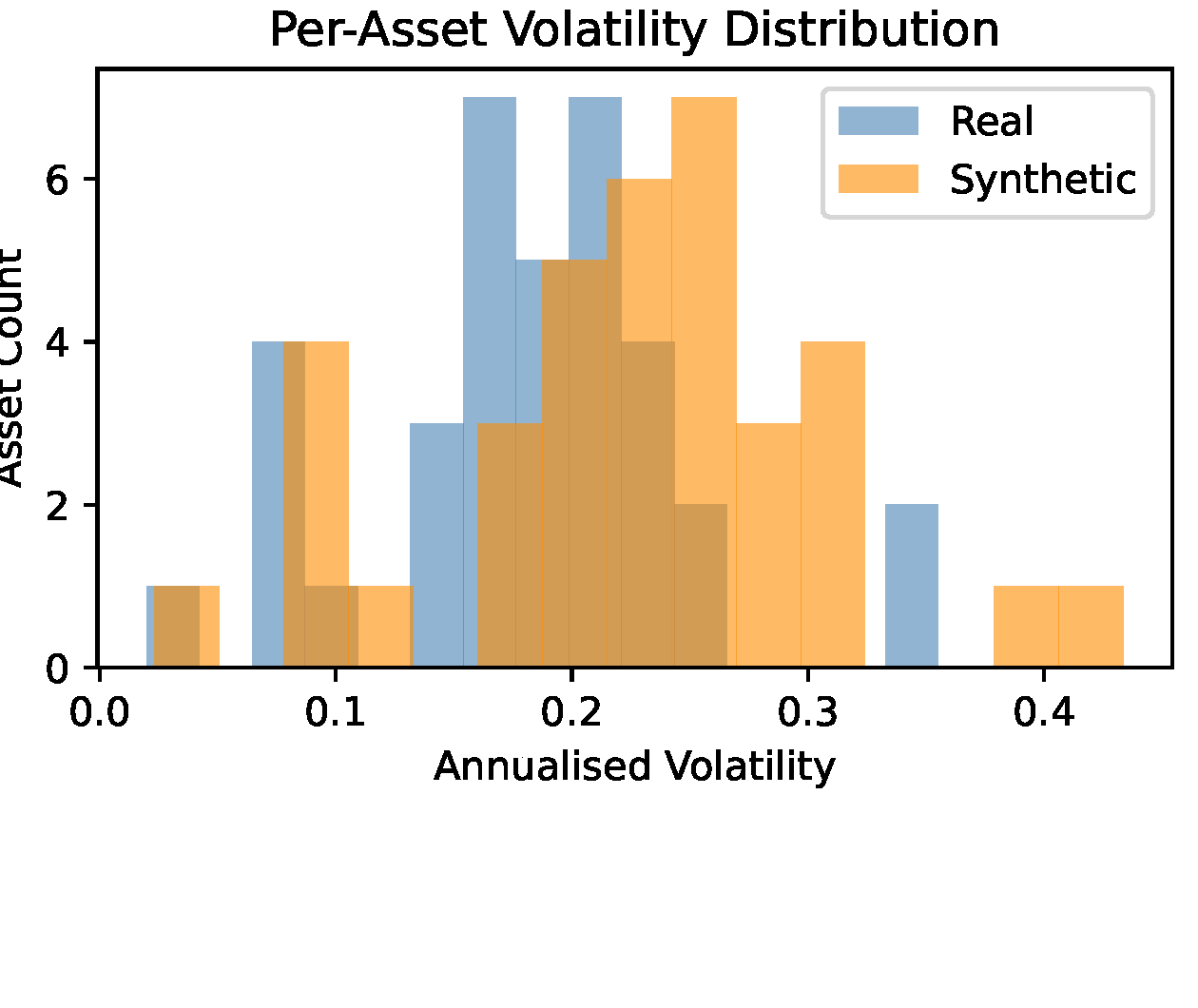}\\[3pt]
    \includegraphics[width=0.98\linewidth]{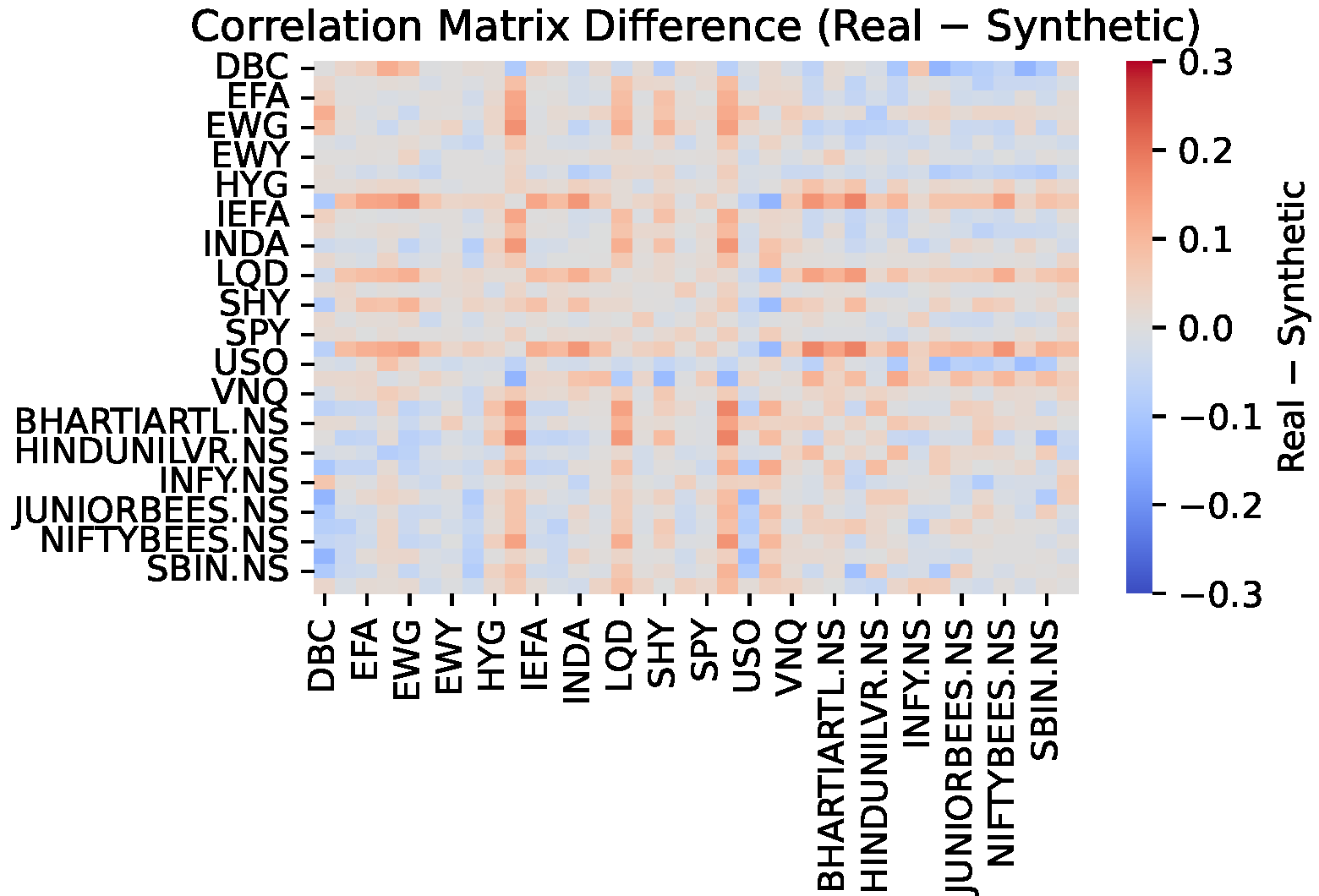}
    \caption{Synthetic data validation. \textit{Top:} Per-asset 
    annualised volatility distribution, real (blue) vs.\ synthetic 
    (orange). Distributions overlap substantially in the 15--25\% 
    range; synthetic returns exhibit a slightly heavier right tail 
    consistent with $t$-copula ($\nu = 6$) calibration. 
    \textit{Bottom:} Element-wise difference between real and 
    synthetic correlation matrices (Real $-$ Synthetic). Most 
    entries fall within $\pm 0.15$; structured deviations in USO 
    and Indian equity names reflect limitations of static factor 
    loadings for regime-dependent assets.}
    \label{fig:synth_valid}
\end{figure}

Figure~\ref{fig:synth_valid} assesses the fidelity of the synthetic 
return panel against real data across two dimensions: marginal 
volatility and cross-asset correlation structure.

\textbf{Marginal volatility.} The per-asset annualised volatility 
distributions (left panel) overlap substantially in the 15--25\% 
range where most assets cluster. Synthetic returns exhibit a slightly 
heavier right tail, consistent with the $t$-copula ($\nu = 6$) 
producing marginally fatter tails than the empirical distribution 
(median $|$weekly return$|$ 1.94\% synthetic vs.\ 1.61\% real).

\textbf{Correlation structure.} The difference heatmap (right panel) 
shows the element-wise gap between real and synthetic correlation 
matrices. Most off-diagonal entries fall within $\pm 0.15$, 
confirming that broad cross-asset dependence is preserved. Two 
structured deviations are visible: USO (crude oil) shows positive 
residuals, indicating that its real correlations with equity assets 
are underestimated by the factor model; and Indian equity names 
show slightly over-estimated mutual correlations in the synthetic 
panel. Both deviations reflect known limitations of static Ridge 
factor loadings for assets with regime-dependent or 
currency-driven co-movement.

These limitations are bounded and expected given the parsimonious 
generation design. The validation is consistent with the synthetic 
pipeline is appropriate for its intended purpose: providing 
structurally plausible training scenarios for the unsupervised 
sandwich phase, not replicating the exact distributional properties 
of the real return panel.

\section{Stress Scenario Definitions}
\label{app:stress_types}

This appendix defines the five stylized stress transformations applied 
to the realized return stream in L2 evaluation. In all cases, the 
decision rule (frozen model weights) is held fixed; only the return 
stream used for realized P\&L is modified, isolating model robustness 
from any adaptive response. Let $\mathbf{X} \in \mathbb{R}^{T \times N}$ 
denote the realized return panel with $T$ weeks and $N$ assets, and let 
$\bar{r}_t = \frac{1}{N}\sum_{i=1}^N X_{ti}$ denote the cross-sectional 
mean return at week $t$.

\subsection{Volatility Bursts}
\label{app:stress_vol_bursts}

Idiosyncratic volatility is amplified during randomly placed burst 
windows. The return panel is decomposed into a common component and 
asset-specific residuals:
\begin{equation}
    X_{ti} = \bar{r}_t + \varepsilon_{ti}, \quad 
    \varepsilon_{ti} = X_{ti} - \bar{r}_t.
\end{equation}
A binary mask $m_t \in \{0,1\}$ is constructed by placing $n_b = 3$ 
non-overlapping burst windows of length $l_b = 8$ weeks at randomly 
selected start points. The stressed panel is:
\begin{equation}
    X^{\text{stress}}_{ti} = \bar{r}_t + 
    \varepsilon_{ti}\bigl(1 + m_t(\sigma_s - 1)\bigr), 
    \quad \sigma_s = 2.0.
\end{equation}
The common component is preserved; only idiosyncratic variance doubles 
during burst periods. This mimics episodes of elevated realized 
volatility without a simultaneous correlation spike.

\subsection{Jumps}
\label{app:stress_jumps}

Market-wide and idiosyncratic jump shocks are superimposed on the 
return panel. At each week $t$, a market jump occurs with probability 
$p_j = 0.03$. Jump magnitudes are drawn from a folded normal: 
$|j_t| \sim |\mathcal{N}(\mu_j, (\mu_j/2)^2)|$ with $\mu_j = 0.08$ 
(8\% weekly). Jump signs are negatively biased with $P(\text{negative}) 
= 0.80$, reflecting the empirical asymmetry of market dislocations. 
The stressed panel is:
\begin{equation}
    X^{\text{stress}}_{ti} = X_{ti} + \mathbf{1}[u_t < p_j] 
    \cdot s_t \cdot |j_t| + \mathbf{1}[v_{ti} < p_j/3] 
    \cdot \eta_{ti},
\end{equation}
where $u_t, v_{ti} \sim \text{Uniform}(0,1)$, $s_t \in \{-1, +1\}$ 
is the jump sign, and $\eta_{ti} \sim \mathcal{N}(0, (\mu_j/2)^2)$ 
are idiosyncratic shocks applied independently at one-third the 
market jump probability.

\subsection{Whipsaw}
\label{app:stress_whipsaw}

Directional reversals are imposed by alternating the sign of the 
market component on successive weeks. Let $a_t = (-1)^t$ be a 
weekly alternating sign sequence. The stressed panel is:
\begin{equation}
    X^{\text{stress}}_{ti} = X_{ti} 
    + \gamma\bigl(a_t |\bar{r}_t| - \bar{r}_t\bigr) 
    - 0.3\gamma\,\varepsilon_{ti}, \quad \gamma = 0.7.
\end{equation}
The first term reverses the market direction at strength $\gamma$; 
the second term partially reverses idiosyncratic residuals. The net 
effect is a mean-reverting, trend-killing environment that penalizes 
momentum-following allocation strategies.

\subsection{Correlation Spike}
\label{app:stress_corr_spike}

All asset returns are blended toward the cross-sectional mean, 
compressing the cross-sectional dispersion:
\begin{equation}
    X^{\text{stress}}_{ti} = (1 - \lambda)\,X_{ti} 
    + \lambda\,\bar{r}_t, \quad \lambda = 0.7.
\end{equation}
At $\lambda = 0.7$, realized pairwise correlations approach unity: 
70\% of each asset's return is driven by the common factor and only 
30\% by asset-specific dynamics. This encodes the empirical observation 
that cross-asset correlations spike toward one during systemic stress 
events, eliminating diversification benefits.

\subsection{Composite Scenario}
\label{app:stress_combo}

The composite scenario applies the three stochastic transformations 
sequentially, using independent random seeds at each stage to avoid 
correlation artifacts:
\begin{equation}
    \mathbf{X}^{(1)} = \text{CorrSpike}(\mathbf{X};\, \lambda=0.7),
\end{equation}
\begin{equation}
    \mathbf{X}^{(2)} = \text{VolBursts}(\mathbf{X}^{(1)};\, 
    \sigma_s=2.0,\, n_b=3,\, l_b=8),
\end{equation}
\begin{equation}
    \mathbf{X}^{\text{stress}} = \text{Jumps}(\mathbf{X}^{(2)};\, 
    p_j=0.03,\, \mu_j=0.08,\, P(\text{neg})=0.80).
\end{equation}
The ordering is deliberate: correlation spike first removes 
diversification, volatility bursts then amplify residual variance, 
and jump shocks finally inject tail events into the already-stressed 
panel. This sequence represents the most adversarial evaluation 
condition in the protocol.


\section{Unlabeled Scenario Window Construction}
\label{app:scenario_windows}

To compute the unsupervised structural loss $\mathcal{L}_{\text{unsup}}$
(Eq.~\ref{eq:unsup_loss}), each unlabeled feature vector must be associated
with a return scenario window. Given the synthetic return panel
$\{\mathbf{r}_t\}$ and unlabeled dates $D_{\text{syn}}$, we form for each
eligible date $t$ a rolling window of length $W$ weeks:
\begin{equation}
\mathbf{R}^{(t)} =
\begin{bmatrix}
\mathbf{r}_{t-W+1}^\top \\
\vdots \\
\mathbf{r}_{t}^\top
\end{bmatrix}
\in \mathbb{R}^{W \times N}.
\label{eq:app_scenario_window}
\end{equation}
Dates without sufficient history (fewer than $W$ preceding synthetic weeks)
are discarded. The resulting scenario matrix $\mathbf{R}^{(t)}$ is used
directly in Eq.~\eqref{eq:scenario_losses} to compute scenario portfolio
losses for the unsupervised CVaR objective.


\section{Deployment Constraint Operator}
\label{app:deployment_constraints}

Training-time objectives do not fully encode all execution and regulatory
constraints. At evaluation we apply a deterministic constraint operator to
map predicted weights $\hat{\mathbf{w}}_t$ to executable weights
$\mathbf{w}^{\text{exec}}_t$.

\subsection{Box Bounds and Renormalization}
\label{app:clip_norm}

Position bounds $w_{\min} \le w_i \le w_{\max}$ (with $w_{\min} = 0$ for
long-only) are enforced by clipping and renormalization:
\begin{align}
\tilde{w}_i &= \min\{w_{\max},\, \max\{w_{\min},\, \hat{w}_i\}\},
\label{eq:app_clip} \\
\mathbf{w}^{(1)} &= \frac{\tilde{\mathbf{w}}}{\|\tilde{\mathbf{w}}\|_1}.
\label{eq:app_renorm}
\end{align}

\subsection{Turnover Cap via Partial Execution}
\label{app:turnover_cap}

Let $\mathbf{w}_{t-1}$ be the previously held weights and $\mathbf{w}^{(1)}_t$
the clipped target. One-way turnover toward the target is:
\begin{equation}
\mathrm{TO}^{\text{target}}_t
= \frac{1}{2}\left\|\mathbf{w}^{(1)}_t - \mathbf{w}_{t-1}\right\|_1.
\label{eq:app_turnover_target}
\end{equation}
If $\mathrm{TO}^{\text{target}}_t > \mathrm{TO}_{\max}$, only a fraction of
the rebalance is executed:
\begin{align}
\alpha_t &= \frac{\mathrm{TO}_{\max}}{\mathrm{TO}^{\text{target}}_t} \in (0,1),
\label{eq:app_alpha_partial} \\
\mathbf{w}^{\text{exec}}_t
&= \mathbf{w}_{t-1}
+ \alpha_t\!\left(\mathbf{w}^{(1)}_t - \mathbf{w}_{t-1}\right).
\label{eq:app_partial_exec}
\end{align}
Clipping and renormalization are re-applied to ensure feasibility.

\subsection{Transaction Cost Model}
\label{app:transaction_costs}

Proportional transaction costs are modeled as a drag on realized returns:
\begin{equation}
\mathrm{TC}_t = c \cdot \mathrm{TO}_t, \quad c > 0,
\label{eq:app_tc}
\end{equation}
where $\mathrm{TO}_t = \frac{1}{2}\|\mathbf{w}^{\text{exec}}_t
- \mathbf{w}^{\text{exec}}_{t-1}\|_1$ is the executed turnover. The net
portfolio return for the first week after a rebalance is:
\begin{equation}
r^{\text{net}}_{p,t} = (\mathbf{w}^{\text{exec}}_t)^\top \mathbf{r}_t
- \mathrm{TC}_t.
\label{eq:app_net_return}
\end{equation}
Transaction costs are not re-applied for subsequent weeks within the same
holding interval, as no rebalance occurs.


\section{C2A Adaptive Inference: Rolling Normalization and Periodic Fine-Tuning}
\label{app:c2a_adaptive}

To adapt to distribution shift on real data while preventing long-horizon
parameter drift, we apply two mechanisms during C2A evaluation: past-only
feature normalization and periodic fine-tuning with reset to the frozen
initialization.

\subsection{Rolling Past-Only Normalization}
\label{app:rolling_norm}

Let $W$ be a rolling window size over decision dates. For each $t$, normalization
statistics are computed strictly from information available before $t$:
\begin{align}
\boldsymbol{\mu}_t &= \frac{1}{W}\sum_{s=t-W}^{t-1} \mathbf{x}_s,
\label{eq:app_roll_mean} \\
\boldsymbol{\sigma}_t &= \sqrt{\frac{1}{W-1}
\sum_{s=t-W}^{t-1}(\mathbf{x}_s - \boldsymbol{\mu}_t)^2},
\label{eq:app_roll_std}
\end{align}
and the normalized feature is:
\begin{equation}
\tilde{\mathbf{x}}_t = \frac{\mathbf{x}_t - \boldsymbol{\mu}_t}
{\boldsymbol{\sigma}_t + \varepsilon}.
\label{eq:app_roll_norm}
\end{equation}
This ensures no future information enters the normalization at any evaluation date.

\subsection{Periodic Fine-Tuning with Turnover Regularization}
\label{app:periodic_finetune}

Every $K$ steps, model parameters are reset to the frozen checkpoint
$\theta_{\text{frozen}}$ and fine-tuned on the most recent $W$ realized periods.
The fine-tuning objective maximizes realized return while discouraging excessive
trading:
\begin{equation}
\begin{split}
\mathcal{L}_{\text{ft}} ={}& -\frac{1}{W}\sum_{s=t-W}^{t-1}\!\left(\mathbf{w}_s^\top \mathbf{r}_{s+1}\right) \\
& + \lambda_{\text{to}} \cdot \frac{1}{W-1}
\sum_{s=t-W+1}^{t-1}\!\left\|\mathbf{w}_s - \mathbf{w}_{s-1}\right\|_1.
\end{split}
\label{eq:app_finetune_loss}
\end{equation}
Resetting to $\theta_{\text{frozen}}$ at each fine-tuning cycle bounds parameter
drift and mitigates overfitting to short windows.


\section{D2A Universe Composition and Transfer Protocol}
\label{app:d2a_universe}

The D2A evaluation universe consists of 36 US-listed instruments organized
into five asset class groups, summarized in Table~\ref{tab:d2a_tickers}.
No Indian equity names are included in D2A, in contrast to the training
universe which contains 12 INR-denominated assets converted to USD. All D2A
assets are quoted directly in USD, eliminating the currency conversion step.

\begin{table}[htbp]
\centering
\caption{D2A out-of-universe evaluation tickers by asset class.}
\label{tab:d2a_tickers}
\begin{adjustbox}{max width=\columnwidth}
\begin{tabular}{llp{6cm}}
\toprule
\textbf{Group} & \textbf{Tickers} & \textbf{Description} \\
\midrule
US Equity factors \& sectors
    & IWM, IWF, IWD, VUG, VTV, MTUM, USMV, XLU, XLE, XLF, XLK, XLV
    & Small-cap, growth, value, momentum, low-volatility, and sector ETFs
      (utilities, energy, financials, technology, healthcare) \\
\addlinespace
International equity
    & VEA, VWO, EWJ, EWG, EWU, INDA, MCHI, EZA
    & Developed markets, emerging markets, and single-country ETFs
      (Japan, Germany, UK, India, China, South Africa) \\
\addlinespace
Fixed income
    & IEF, TLT, SHY, LQD, HYG, TIP
    & Treasury (short, intermediate, long), investment-grade credit,
      high-yield credit, inflation-protected \\
\addlinespace
Real assets
    & GLD, SLV, DBC, USO, UNG, VNQ
    & Gold, silver, broad commodities, oil, natural gas, REITs \\
\addlinespace
Alternatives / FX
    & UUP, FXE, FXY, GSG
    & USD index, EUR/USD, JPY/USD, broad commodity index \\
\bottomrule
\end{tabular}
\end{adjustbox}
\end{table}

\paragraph{Overlap with training universe.}
The training universe (C2A) consists of 23 US ETFs and 12 Indian equities.
Structural overlap with D2A arises through shared broad-market and fixed-income
exposure: instruments such as TLT, GLD, and broad equity proxies appear in both
universes or have close functional equivalents. This partial continuity
(approximately 40\% by structural role) ensures that core risk-reduction
heuristics learned during training --- such as flight to fixed income under
stress or commodity diversification --- remain applicable in D2A, while the
factor ETFs and sector instruments (MTUM, USMV, XLU, XLV etc.) constitute
genuinely unseen allocation targets.

\paragraph{Feature pipeline parameters.}
The same feature schema described in Appendix~\ref{app:data_details} is applied
to D2A assets without modification. Pipeline parameters are: rolling ridge
lookback 52 weeks, ridge regularization $\alpha = 5.0$, mean factor horizon
13 weeks, minimum observations per regression 30, and blending weights
$w_{\mu} = 0.7$, $w_{\text{mom}} = 0.3$. Models are evaluated on D2A.
D2A evaluation follows the same adaptation protocol as C2A, with periodic fine-tuning and reset, but applied to a disjoint asset universe. 


\section{Additional Experimental Results}
\label{app:additional_results}

This appendix provides supplementary figures for the GRID\_3$\times$5
evaluation discussed in Section~\ref{subsec:grid_results}, including
per-seed Sharpe distributions for world seeds 32 and 52, the cumulative
wealth trajectory for seed 32, the CVaR distribution across all runs,
the mean Sharpe confidence interval plot, and the stability win-rate
heatmap. These figures support the regime dependency findings summarized
in Section~\ref{subsec:bayesian_vs_det}.

\subsection{Per-Seed Sharpe Distributions}
\label{app:per_seed_sharpe}

\begin{figure}[!htbp]
    \centering
    \includegraphics[width=0.8\linewidth]{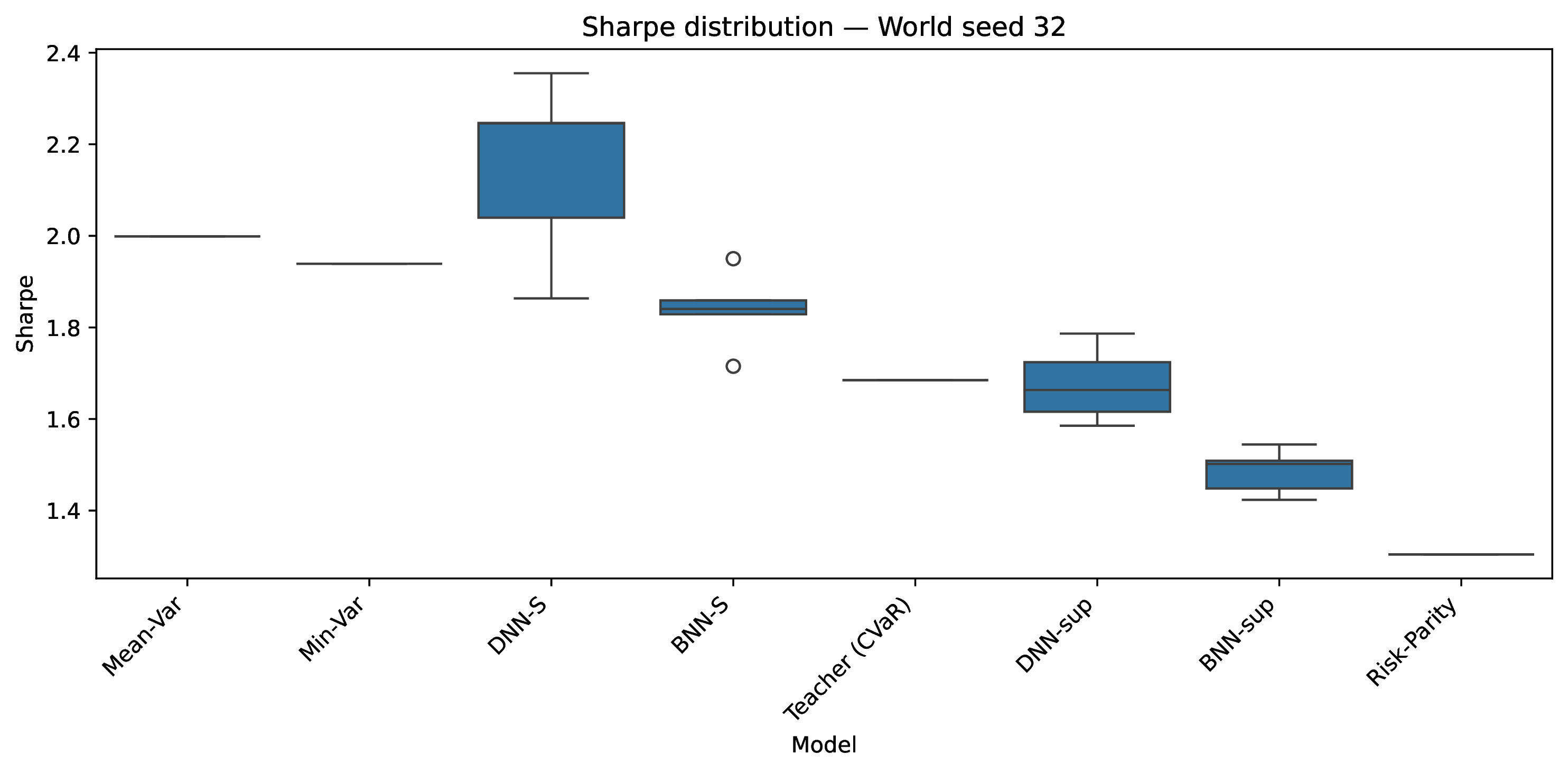}
    \caption{Sharpe distributions under seed 32 (bull market regime,
    2016--2018 tech expansion). DNN-S achieves peak performance (Sharpe
    2.15), outpacing Mean-Variance (1.99) and BNN-S (1.83). Momentum
    strategies thrive in persistent trending environments, giving
    deterministic overweighting of high-momentum assets a structural
    advantage. Risk-Parity trails at 1.30.}
    \label{fig:Seed32SharpeDist}
\end{figure}

\begin{figure}[!htbp]
    \centering
    \includegraphics[width=0.8\linewidth]{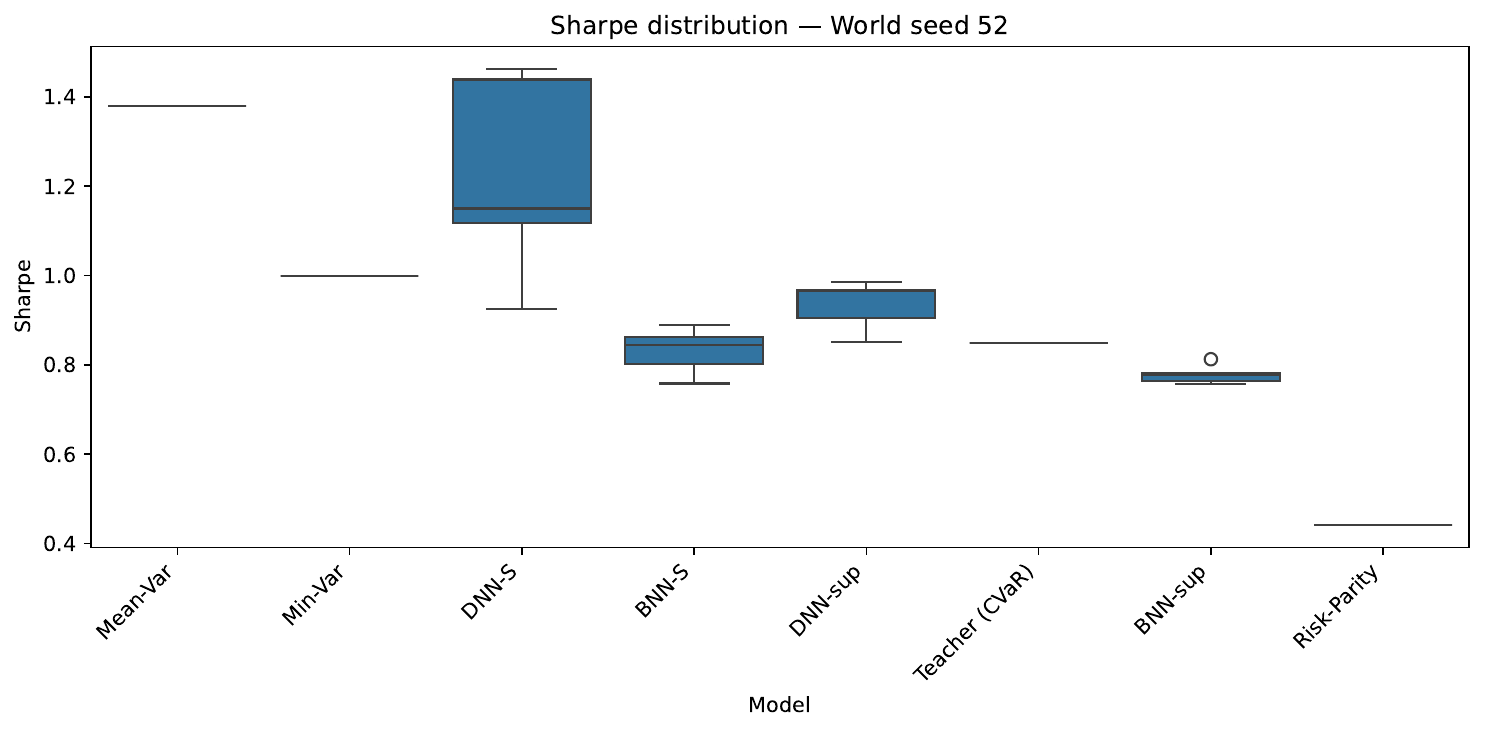}
    \caption{Sharpe distributions under seed 52 (mixed regime). Mean-Variance
    recovers partially (1.38) relative to seed 42, and DNN-S leads among ML
    models (1.22). Risk-Parity exhibits catastrophic failure (0.45,
    $-66\%$ relative to seed 32), confirming that equal-risk-weighting is
    particularly fragile under rapid regime transitions.}
    \label{fig:Seed52SharpeDist}
\end{figure}

Figures~\ref{fig:Seed32SharpeDist} and~\ref{fig:Seed52SharpeDist} show
Sharpe ratio distributions for world seeds 32 and 52 respectively, which
represent a bull market regime and a mixed regime. Together with seed 42
(Figure~\ref{fig:Seed42SharpeDist} in the main text), these illustrate the
full range of regime-dependent behavior across the three world seeds.

\FloatBarrier

\FloatBarrier

\subsection{Cumulative Wealth Trajectory}
\label{app:equity_curve}

\begin{figure}[!htbp]
    \centering
    \includegraphics[width=0.8\linewidth]{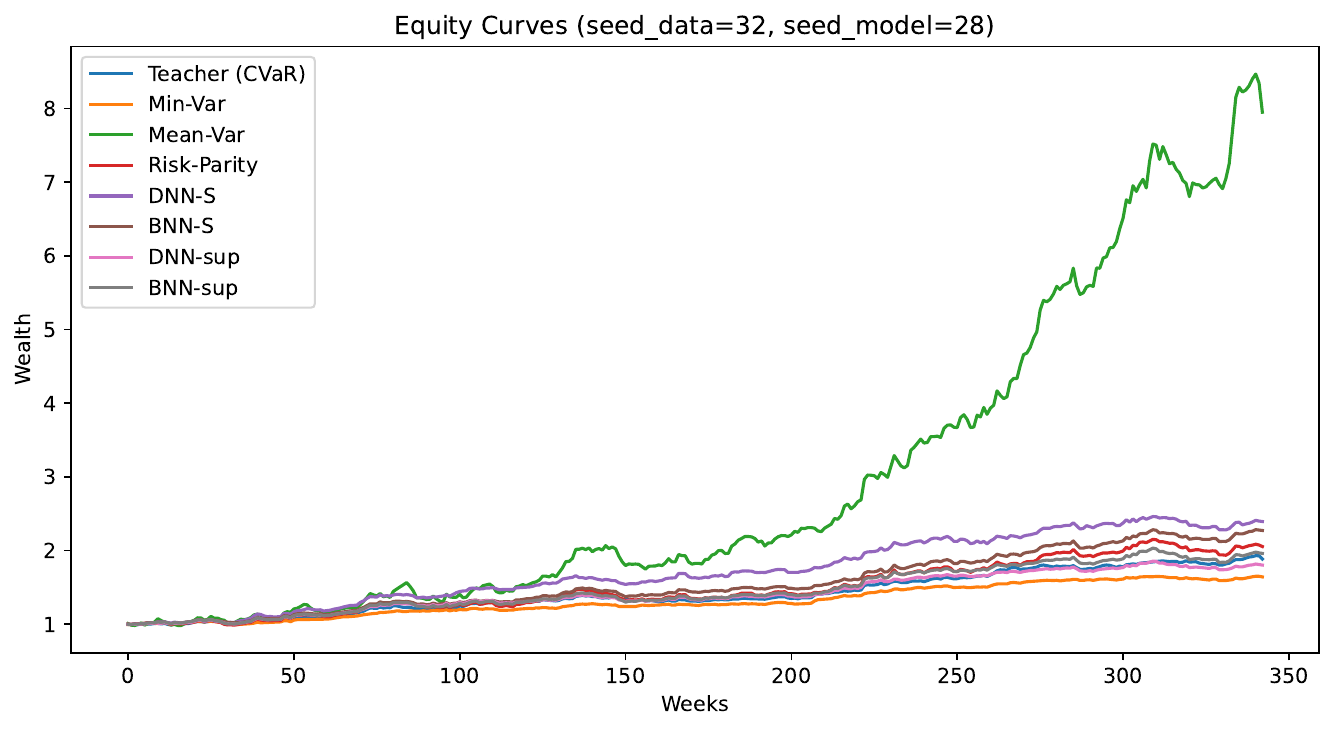}
    \caption{Cumulative wealth over $\sim$350 weeks (seed 32). Mean-Variance
    (green) exhibits exponential growth after week 200 ($9\times$ terminal
    wealth), capturing bull market momentum through concentrated positions.
    ML models (BNN-S, DNN-S) maintain steady $2.3\times$ growth with
    substantially lower drawdowns. The divergence at week 200 is consistent
    with a regime shift from balanced to trending markets.}
    \label{fig:Seed32EquityCurve}
\end{figure}

Figure~\ref{fig:Seed32EquityCurve} shows the cumulative wealth evolution
over approximately 350 weeks (7 years) under seed 32. The divergence between
Mean-Variance and ML models after week 200 is consistent with a regime shift
toward persistent momentum, which disproportionately benefits concentrated
mean-variance allocations while BNN-S and DNN-S maintain steadier compounding
with lower drawdowns.

\FloatBarrier

\subsection{CVaR Distribution and Confidence Interval Plots}
\label{app:cvar_ci_plots}

\begin{figure}[!htbp]
    \centering
    \includegraphics[width=0.8\linewidth]{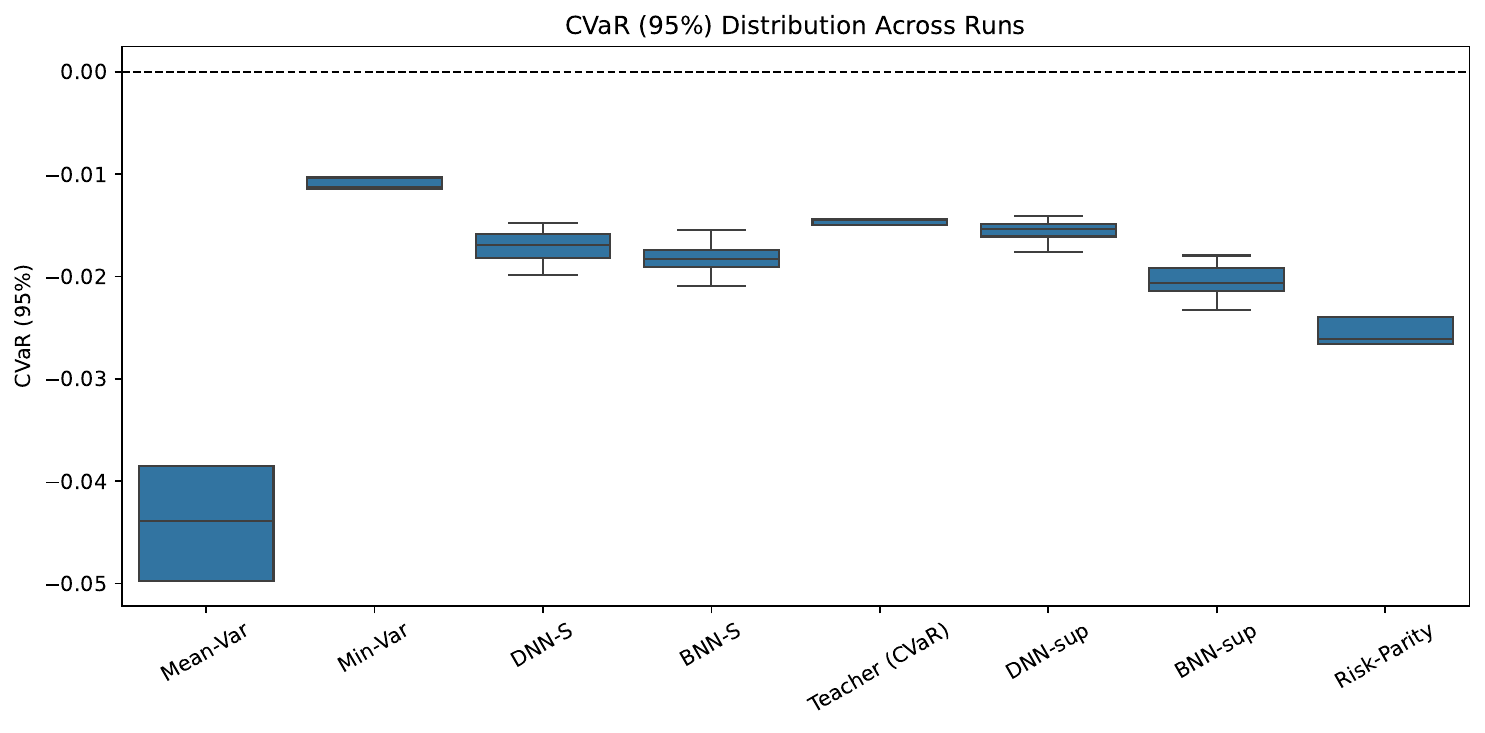}
    \caption{Distribution of CVaR (95\%) expressed as weekly loss percentage
    across 15 runs per model (GRID\_3$\times$5). Lower values indicate
    worse tail risk. Mean-Variance exhibits median CVaR of $-4.5\%$ with
    wide dispersion, approximately $2.4\times$ worse than BNN-S ($-1.8\%$).
    BNN models show tighter distributions ($\pm 0.3\%$) than DNN variants
    ($\pm 1.0\%$), consistent with epistemic uncertainty reducing
    overconfident extreme positions \citep{Blundell2015BayesByBackprop}.}
    \label{fig:CVaRacrossRuns}
\end{figure}

\begin{figure}[!htbp]
    \centering
    \includegraphics[width=0.8\linewidth]{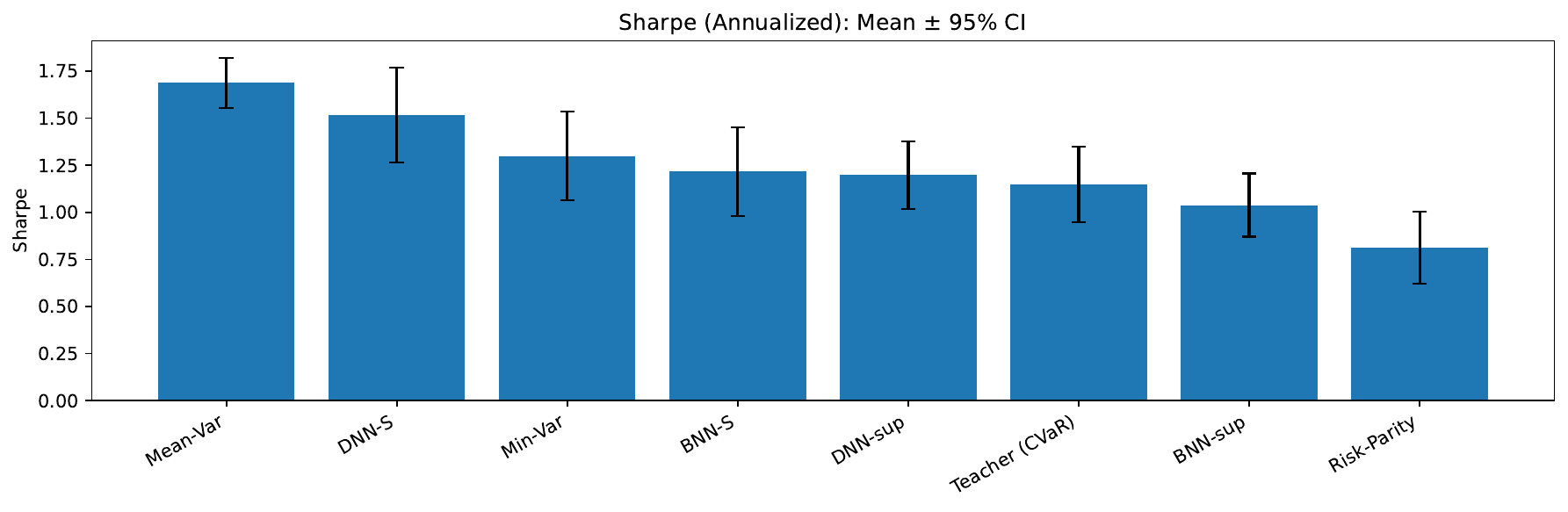}
    \caption{Mean Sharpe ratios with 95\% bootstrap confidence intervals
    ($n = 1000$ resamples). Bootstrap confidence intervals show BNN-S with
    higher mean Sharpe than the Teacher, though intervals show partial
    overlap given the small sample ($n=15$).
    Mean-Variance confidence interval ($[1.56, 1.82]$) overlaps with
    BNN-S ($[0.98, 1.45]$), and its tail-risk disqualification
    (Figure~\ref{fig:CVaRacrossRuns}) renders raw Sharpe superiority
    misleading for risk-averse deployment.}
    \label{fig:Sharpe_Annualized_Mean_PM_95_CI}
\end{figure}

Figure~\ref{fig:CVaRacrossRuns} shows the distribution of CVaR (95\%)
across all 15 runs per model, complementing the Sharpe distributions in
Figure~\ref{fig:sharp_dist_across_runs}. Figure~\ref{fig:Sharpe_Annualized_Mean_PM_95_CI}
reports mean Sharpe ratios with 95\% bootstrap confidence intervals,
providing the statistical basis for the significance claims in
Section~\ref{subsec:grid_results}.

\FloatBarrier

\FloatBarrier

\subsection{Stability Win-Rate Heatmap}
\label{app:stability_heatmap}

\begin{figure}[!htbp]
    \centering
    \includegraphics[width=0.8\linewidth]{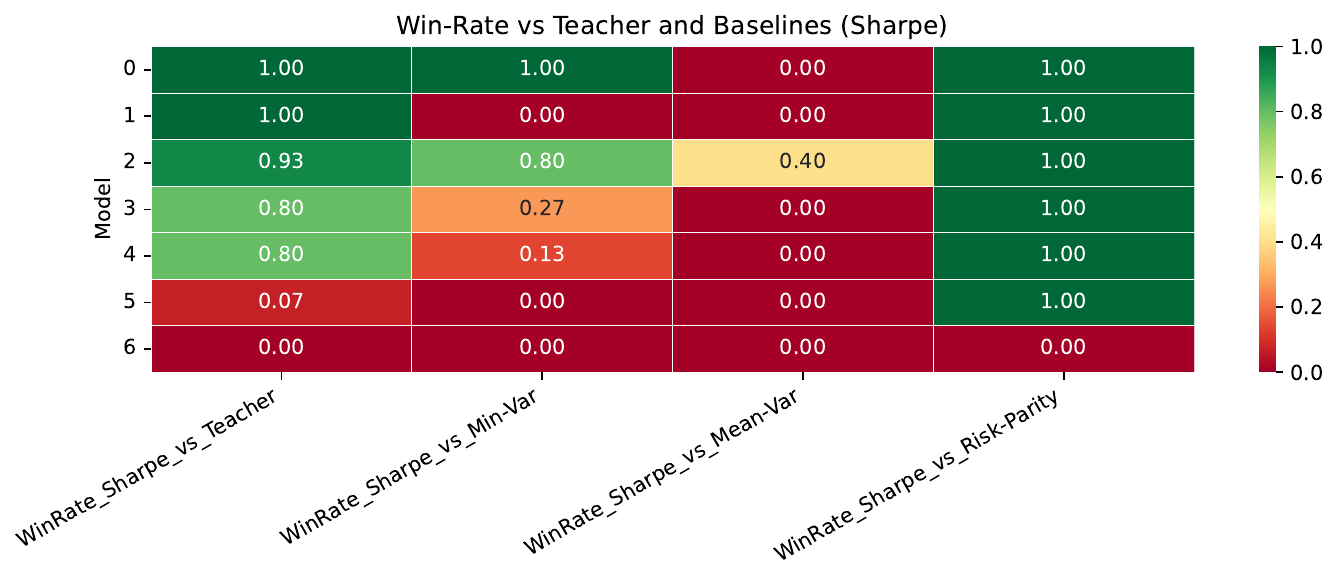}
    \caption{Stacked win-rate heatmap across all model pairs
    (GRID\_3$\times$5, 15 runs). Each row represents one model, columns
    represent opponents, and color intensity indicates win percentage.
    Model 0 corresponds to Mean-Variance and achieves 100\% win-rates
    against Teacher, Risk-Parity, and BNN-S on raw Sharpe. By contrast,
    BNN-S (row 3) records 80\% against Teacher and 27\% against
    Min-Variance and 0\% against Mean-Var. The raw-Sharpe dominance of Mean-Variance must be
    interpreted alongside tail-risk analysis (Figure~\ref{fig:CVaRacrossRuns}),
    which reveals Mean-Variance's
    Pareto-dominated position for risk-averse investors
    \citep{rockafellar2000}.}
    \label{fig:StabilityHeat}
\end{figure}

Figure~\ref{fig:StabilityHeat} provides a stacked win-rate comparison
across all model pairs, complementing the pairwise dominance matrix
(Figure~\ref{fig:PairwiseMatrix}) in the main text. Color intensity
indicates win percentage for each row model against each column opponent.

\FloatBarrier

\section*{Generative AI Disclosure}
Generative AI tools were used during the preparation of this manuscript for language
refinement, formatting, and structural assistance. All scientific content,
methodology, experiments, interpretations, and final verification were
carried out by the author.

\bibliography{references}

\end{document}